\documentclass{article} % For LaTeX2e
 \PassOptionsToPackage{numbers,sort&compress}{natbib}
 \usepackage[preprint]{neurips_2026}
\usepackage{amsmath,amsfonts,bm}

\def\eqref#1{equation~\ref{#1}}
\def\1{\bm{1}}

\DeclareMathAlphabet{\mathsfit}{\encodingdefault}{\sfdefault}{m}{sl}
\SetMathAlphabet{\mathsfit}{bold}{\encodingdefault}{\sfdefault}{bx}{n}

\newcommand{\R}{\mathbb{R}}

\DeclareMathOperator*{\argmin}{arg\,min}

\usepackage[utf8]{inputenc} % allow utf-8 input
\usepackage[T1]{fontenc}    % use 8-bit T1 fonts
\usepackage{url}            % simple URL typesetting
\usepackage{booktabs}       % professional-quality tables
\usepackage{amsfonts}       % blackboard math symbols
\usepackage{nicefrac}       % compact symbols for 1/2, etc.
\usepackage{microtype}      % microtypography
\usepackage{xcolor}         % colors
\usepackage{graphicx}
\usepackage{grffile}
\usepackage{subcaption}
\usepackage{pifont}         % for \ding
\usepackage{adjustbox}
\usepackage{multirow}
\usepackage{makecell}
\usepackage{float}
\usepackage{algorithm}
\usepackage{algorithmic}
\usepackage{amsthm}
\theoremstyle{definition} % Optional: sets font style to upright instead of italic
\newtheorem{observation}{Observation}
\usepackage{amsmath,amssymb,mathtools,bm,empheq,amsthm}

\usepackage{tikz}
\usetikzlibrary{positioning,arrows.meta,shapes.geometric,calc,matrix,backgrounds,fit,plotmarks}
\usepackage{pgfplots}
\pgfplotsset{compat=1.18}
\usepgfplotslibrary{groupplots}

\usepackage[textsize=tiny]{todonotes}

\usepackage{hyperref}
\usepackage[capitalize,noabbrev]{cleveref}

\newcommand{\m}[1]{\mathbf{#1}}    % For bold matrices like \m{W}
\title{ROCKET: Rapid Optimization via Calibration-guided Knapsack-Enhanced Truncation for Efficient Model Compression}

\author{%
  \textbf{Ammar Ali$^{1,2*}$,}
  \textbf{Baher Mohammad$^{1,2*}$,}
  \textbf{Denis Makhov$^{1,2}$,} \\
  \textbf{Dmitriy Shopkhoev$^{1}$,} 
  \textbf{Magauiya Zhussip$^{1,2}$, }
  \textbf{Stamatios Lefkimmiatis$^{1,2}$} \\
  $^{1}$MWS AI $\quad$
  $^{2}$ITMO \\
  \texttt{ammarali32@itmo.ru}\\
}

\theoremstyle{plain}
\newtheorem{theorem}{Theorem}[section]
\newtheorem{proposition}[theorem]{Proposition}
\newtheorem{lemma}[theorem]{Lemma}

\theoremstyle{definition}

\theoremstyle{remark}

\def\bal#1\eal{\begin{align}#1\end{align}} % align environment

\newcommand{\pr}[1]{\left(#1\right)} % left-right parenthesis 
\def\m{\mathbf}
\def\mc{\mathcal}
\def\R{\mathbb{R}}

\newcommand{\norm}[2]{\ensuremath{\left\|#1\right\|_{#2}}}

\newcommand {\bbmtx}{\begin{bmatrix}} % begin matrix environment
\newcommand {\ebmtx}{\end{bmatrix}} % end matrix environment

\DeclareMathOperator*{\trace}{tr} % diagonal
\newcommand{\tr}[1]{\trace\pr{#1}}

\newcommand{\ours}{ROCKET}
\newcommand{\cmark}{\ding{51}}%
\newcommand{\xmark}{\ding{55}}%

\begin{document}

\maketitle
% equal-contribution footnote (marks are set manually in the author block)
\begingroup
\renewcommand{\thefootnote}{\fnsymbol{footnote}}
\footnotetext[1]{Equal contribution.}
\endgroup

\begin{abstract}
We present \textbf{ROCKET}, a training-free model compression method that achieves state-of-the-art performance compared with factorization, structured-sparsification, and dynamic compression baselines. Operating under a global compression budget, ROCKET comprises two key innovations. First, it formulates layer-wise compression allocation as a multi-choice knapsack problem, selecting the best compression configuration for each layer to minimize total reconstruction error while adhering to a target model size. Second, it introduces a single-pass sparse matrix factorization that executes what is, conceptually, one iteration of the classical alternating dictionary-learning template with both half-steps replaced by closed-form, calibration-aware computations: a spectral initialization followed by importance-weighted hard thresholding that accounts for the metric distortion between whitened and original weight spaces, and a closed-form dictionary update via least squares, bypassing iterative optimization, sparse coding, or backpropagation entirely. \ours{} consistently outperforms existing compression approaches across different model architectures at 20--50\% compression rates. Notably, it retains over 90\% of the model's performance at 30\% compression without any fine-tuning. Code implementing ROCKET is available at \url{https://github.com/mts-ai/ROCKET/tree/main}.
\end{abstract}
\section{Introduction}

Transformers have achieved unprecedented success across vision and language tasks, but modern LLMs now scale to billions and trillions of parameters, imposing prohibitive computational and memory demands that hinder deployment on edge devices and latency-sensitive applications. To address this, research focuses on three primary compression strategies: pruning \cite{pruning}, distillation \cite{distillation}, and post-training weight factorization \cite{wang2025svdllm}. Among these, post-training weight factorization has emerged as a particularly promising direction, enabling substantial parameter reduction without the need for costly retraining or fine-tuning. A dominant paradigm in this area is low-rank approximation using truncated Singular Value Decomposition (SVD), which approximates each weight matrix as the product of two smaller dense matrices.  However, this strategy imposes a rigid structural constraint forcing all columns of the weight matrix to lie in a single shared low-dimensional subspace. This often limits the representational capacity and leads to significant performance degradation under moderate to high compression.
This limitation has spurred the development of methods that go beyond a single shared subspace representation, adopting instead a union-of-subspaces framework akin to dictionary learning. In such models, a weight matrix is expressed as a combination of a subset of basis matrices \cite{zhussip2025shareattention}, or alternatively, its individual columns are represented as sparse linear combinations of atoms from a shared dictionary~\cite{Shopkhoev2025cospadi0}. These formulations provide greater flexibility by capturing the heterogeneous local structures present within the weight matrix. 
Despite their theoretical appeal, practical adoption of these methods faces severe computational challenges: conventional dictionary learning algorithms rely on iterative alternating minimization between sparse coding and dictionary update steps~\cite{aharon2006ksvd}, which do not scale well for large-scale LLM weight matrices.

In this work, we propose \textbf{\ours}, a fast, training-free compression method that addresses  the representational rigidity of low-rank factorization while avoiding the computational burden of iterative dictionary learning. Our approach introduces two key contributions. First, \ours{} compresses weight matrices via what is, conceptually, a \textbf{single iteration of the classical alternating dictionary-learning template}, in which both half-steps are replaced by closed-form, calibration-aware computations. This yields a factorization that inherits the expressive power of union-of-subspaces models yet operates substantially faster than alternating minimization. Second, rather than applying uniform compression or relying on heuristic layer-wise sensitivity estimates, \ours{} formulates global compression allocation as a multi-choice knapsack problem. For each layer, it selects the best compression configuration from a set of precomputed candidates to minimize total weight reconstruction error under a target model size. Together, these components enable \ours{} to produce compact models that achieve substantially higher accuracy compared to existing post-training compression methods.

The contributions of this work are summarized as follows:
(1) We propose \ours{}, an efficient, training-free LLM compression method that factorizes weight matrices into a sparse dictionary representation computable in a single closed-form pass, without iterative optimization;
(2) We introduce a calibration-guided criterion for sparsifying the coefficient matrix that accounts for the metric distortion between original and whitened weight spaces to preserve salient directional information;
(3) We formulate layer-wise compression allocation as a multi-choice knapsack problem, enabling dynamic, performance-aware distribution of the global compression budget across layers;
(4) Through extensive experiments, we demonstrate that \ours{} consistently outperforms state-of-the-art compression methods including structured sparsification, low-rank factorization, and adaptive budget allocation techniques across multiple modalities (text, vision, and audio).\vspace{-.4cm}

\section{Related Work}\vspace{-.3cm}
This work intersects three primary research directions in model compression: (1) dynamic per-layer allocation of compression budgets, (2) structured matrix factorization for weight approximation, and (3) sparsification techniques. We review recent advances in each area, with emphasis on methods most relevant to our training-free, reconstruction-aware compression framework.

\textbf{Structured Matrix Factorization for Weight Approximation}
Early approaches employed truncated SVD to obtain low-rank approximations of transformer weights. However, several studies \cite{ASVD, wang2025svdllm, drone} demonstrated that weight matrices themselves are not inherently low-rank; instead, activations exhibit low-rank structure. These works proposed data-aware low-rank approximations using a whitening transform estimated from a small calibration dataset, yielding significantly more effective compression.
A more general representation was recently introduced in \cite{Shopkhoev2025cospadi0}, where weights are expressed as sparse linear combinations of dictionary atoms in a whitened space, computed via K-SVD and Orthogonal Matching Pursuit (OMP) updates. This approach overcomes the limitation of fixed, layer-invariant bases in low-rank methods since it allows each column of the weight matrix to reside in a different low-dimensional subspace, effectively promoting a more flexible union-of-subspaces modeling strategy.
% Our approach extends this line of work by executing what is, conceptually, a single iteration of the classical alternating dictionary-learning template~\cite{engan1999method, aharon2006ksvd}, with both half-steps replaced by closed-form, calibration-aware computations. This eliminates iterative optimization while achieving higher reconstruction accuracy and substantially faster compression (930\,s vs.\ 90,080\,s for Llama3.2-1B), which is critical for scaling to billion-parameter models.
ROCKET bridges SVD efficiency and dictionary-learning expressiveness~\cite{engan1999method, aharon2006ksvd} through a single closed-form, calibration-aware approximation.

\textbf{Budget Allocation and Layer Importance}
Many early compression methods apply uniform compression across all layers, implicitly assuming equal layer importance. Recent work challenges this assumption. LLM-Pruner \cite{llmpruner} estimates the importance of coupled layer groups using gradient and Hessian-based metrics, pruning less critical groups. ARS \cite{ars} proposes an adaptive rank selection mechanism using differentiable binary masks, regularized to respect the ordering of singular values from SVD, thereby allocating more capacity to important layers.
Dobi-SVD \cite{dobisvd} introduces a learnable truncation threshold $k$ per weight matrix, optimized during training via a multi-objective loss balancing task performance and global compression ratio. Similarly, ARA \cite{ara} dynamically assigns ranks to linear modules by learning a monotonic probabilistic mask over singular values, guided by a loss that accounts for cases where full-rank retention is more efficient than decomposition.
Hardware-aware pruning methods such as HALP~\cite{shen2021halp} also formulate compression as a constrained selection problem, using latency estimates and saliency scores to choose pruned structures. These methods are typically training-aware and operate on binary prune/retain decisions over channels or filters.
In contrast, our method performs allocation in a purely post-training setting via a multi-choice knapsack formulation. Using dynamic programming (DP), we select the best configuration over the evaluated candidates minimizing total reconstruction error under a per-layer error cap. Unlike prior DP approaches for CNN sparsity \cite{FrantarA22} that rely on heuristic sensitivity scores, ROCKET uses an analytical, calibration-guided error from activation-aware eigendecomposition. Our dual-constraint formulation maps to an exact shortest-path problem on a Directed Acyclic Graph (DAG) with dominated-state pruning, yielding reduced computational cost in practice.

% \textbf{Sparsification Methods}
% Unstructured pruning has demonstrated strong efficacy in compressing large language models. While magnitude-based pruning~\cite{magnitudesparse} remains a simple baseline, Frantar et al.~\cite{frantar2023sparsegpt} showed its inadequacy for LLMs and proposed SparseGPT, a Hessian-aware, layer-wise pruning method that reconstructs output errors via an efficient approximate solver.
% Alternative importance metrics have also been explored. WANDA~\cite{wanda} computes a saliency score as the product of weight magnitude and the  $\text{L}_2$   norm of corresponding input activations (estimated from calibration data), pruning the lowest-scoring weights per output neuron. Bonsai~\cite{bonsai} formulates module importance as an underdetermined regression problem, estimating importance using only forward passes to enable efficient structured pruning.
% Although sparsification achieves high compression ratios, it often yields irregular memory access patterns that hinder inference acceleration on modern hardware.
%% recent works show that it is possible to get a speedup using Macko it is a direction with a lot of interest 
\textbf{Sparsification Methods}
We emphasize that \ours{} is not a pruning method: sparsity is applied to the coefficient matrix factor of the weights, not to their original entries, and the resulting product $\m{F}\m{H}$ is preserved for inference. While
structured sparse factorizations offer native hardware compatibility, recent studies show that unstructured sparsity starting at roughly 30\% can also deliver practical speedups \cite{macko}, highlighting a growing area of interest.\vspace{-0.3cm}
\section{Method}
\vspace{-0.3cm}
\label{sec:method}

\begin{figure*}[t]
    \centering
    \includegraphics[width=0.95\linewidth]{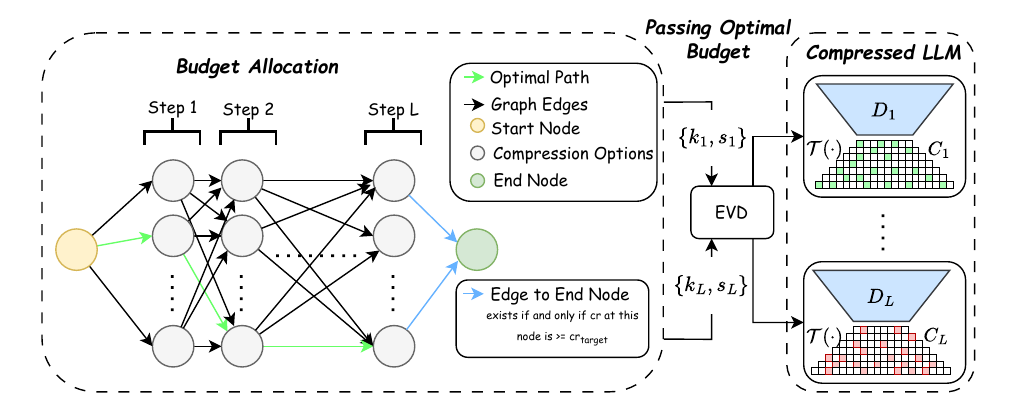}
    \caption{
        Overview of the proposed method.
        \textbf{Left:} Budget allocation formulated as a shortest-path problem on a directed
        graph, where nodes represent compression options and edges encode cost
        (reconstruction error), solved via dynamic programming to find the best
        sequence of operations over the discretized candidate set.
        \textbf{Right:} The selected path determines per-layer compression
        parameters (rank $k_\ell$ and sparsity $s_\ell$), which are then applied to each
        layer via eigendecomposition followed by structured hard thresholding
        sparsification ($\mathcal{T}(\cdot)$) of coefficients.
    }
    \label{mainmethod}\vspace{-.5cm}
\end{figure*}

Recent training-free compression methods for LLMs exhibit a fundamental trade-off between computational efficiency and representational flexibility. On one end, truncated SVD (SVD-LLM;~\cite{wang2025svdllm}) enforces a
rigid, globally shared low-rank subspace where compression is fast, but reconstruction fidelity degrades severely under aggressive ratios. On the other end, CoSpaDi~\cite{Shopkhoev2025cospadi0} employs sparse
dictionary learning (K-SVD + OMP~\cite{aharon2006ksvd}) to realize a union-of-subspaces model, at the cost of a runtime that scales poorly to multi-billion-parameter models (646~min for Llama3.2-1B).
\ours{} bridges this gap through the half-step replacement with the closed-form updates; and its allocation stage distributes a global parameter budget across layers by solving a constrained knapsack problem
over the profiled candidates (Sections~\ref{sec:profiling}--\ref{sec:dp}).

\subsection{Problem Setup and Whitened Activation Space}
\label{sec:whitening}

Given a calibration dataset $\m{X}\in\R^{N\times d_1}$ and a weight matrix $\m{W}\in\R^{d_1\times d_2}$, we seek an approximation $\widetilde{\m{W}}$ that minimizes the activation-aware reconstruction error
\bal
\argmin_{\widetilde{\m{W}}\in \mc{C}}\;
  \norm{\m{X}\m{W}-\m{X}\widetilde{\m{W}}}{F}\equiv\tr{\pr{\m W-\widetilde{\m W}}^\top\m X^\top\m X\pr{\m W-\widetilde{\m W}}}^{\frac{1}{2}}
  %\quad\text{s.t.}\quad
  %\widetilde{\m{W}}\in\mc{C},
\eal
where $\mc{C}$ imposes structural constraints on $\widetilde{\m{W}}$.
Following standard practices in data-aware compression, we consider the regularized Gram matrix $\m{A}=\m{X}^{\top}\m{X}+\epsilon\m{I}$ ($\epsilon=10^{-6}$), to ensure that it is positive definite, and we define $\m L$ to be its Cholesky factor, that is $\m A =\m L^\top\m L$. Then the original objective above can be equivalently expressed as $\norm{\m{W}_L-\widetilde{\m{W}}_L}{F}$ with
$\widetilde{\m{W}}_L=\m{L}\widetilde{\m{W}}$. This problem reformulation allows us to work directly on the whitened activation space, which encodes all necessary information from the calibration data.

% established data-aware compression
% strategies~\cite{chen2021drone,yuan2023asvd,wang2025svdllm}, we operate in the \emph{whitened activation space}.  Let $\m{L}$ be the Cholesky factor of the Gram matrix
% $\m{A}=\m{X}^{\top}\m{X}+\epsilon\m{I}$ ($\epsilon=10^{-6}$), and define the decorrelated input
% $\m{Y}=\m{X}\m{L}^{-1}$, which satisfies $\m{Y}^{\top}\m{Y}=\m{I}_{d_1}$. Then
% \begin{equation}
%   \norm{\m{X}\m{W}-\m{X}\widetilde{\m{W}}}{F}
%   =\norm{\m{L}\m{W}-\m{L}\widetilde{\m{W}}}{F},
%   \label{eq:whitening}
% \end{equation}
% so with the whitened weight $\m{W}_L:=\m{L}\m{W}$ the objective reduces to
% $\norm{\m{W}_L-\widetilde{\m{W}}_L}{F}$ with
% $\widetilde{\m{W}}_L=\m{L}\widetilde{\m{W}}$: all
% activation-awareness is absorbed into the geometry of the whitened space.

\subsection{Dictionary Learning without the Loop}
\label{sec:dl-view}

In the whitened space the structural constraint takes the form of a
sparse dictionary factorization
$\widetilde{\m{W}}_L=\m{D}\m{S}$:
\begin{equation}
  \min_{\m{D}\in\R^{d_1\times k},\;\m{S}\in\R^{k\times d_2}}
  \norm{\m{W}_L-\m{D}\m{S}}{F}^{2}
  \quad\text{s.t.}\quad
  \m{S}_{:j}\text{ is $s$-sparse for every column }j.
  \label{eq:dl}
\end{equation}
The joint problem~\eqref{eq:dl} is non-convex and is classically solved by \emph{alternating minimization} between (a)~a sparse coding step (updating $\m{S}$ for fixed $\m{D}$), attacked greedily by OMP-type
pursuit~\cite{tropp2007signal}, and (b)~a dictionary update (updating $\m{D}$ for fixed $\m{S}$), either via the Method of Optimal Directions (MOD~\cite{engan1999method}) or by
K-SVD~\cite{aharon2006ksvd}.  Both half-steps are expensive and many alternations are required; this is precisely what limits the scalability of CoSpaDi~\cite{Shopkhoev2025cospadi0}.

\ours{} is built on a simple observation:
\begin{observation}[Column-orthogonal dictionaries make sparse-coding half-step closed-form]
\label{obs:orthogonal}
If $\m{D}^{\top}\m{D}=\m{I}_k$, then (a)~the sparse coding step of~\eqref{eq:dl} decouples across columns and is solved by hard thresholding (Appendix~\ref{app:pythagoras}), and (b)~dropping the imposed orthogonality constraint, the dictionary
update is a ridge least-squares problem with a closed-form solution (Section~\ref{sec:refinement}).
\end{observation}

\ours{} therefore executes \emph{exactly one} iteration of the alternating template, with each half-step replaced by its closed-form, calibration-aware counterpart:

  [\textbf{S1}] \emph{Spectral initialization:} the optimal column-orthogonal dictionary for the requested rank is computed, with the coefficients obtained as orthogonal projections of the whitened weights.
  (Section~\ref{sec:spectral});

  [\textbf{S2}] \emph{Calibrated hard thresholding:} An importance-weighted thresholding rule that accounts for the un-whitening distortion is applied on the coefficients. (Section~\ref{sec:thresholding});
  
  [\textbf{S3}] \emph{Dictionary refinement:} MOD dictionary update is performed with the orthogonality constraint being dropped; this constraint was initially imposed solely to simplify the sparse-coding step and it does not represent a modeling requirement (Section~\ref{sec:refinement}).

The method deliberately terminates after this single pass: step~S3 destroys the column-orthogonality that made step~S2 a closed-form solver, so any subsequent sparse-coding step would require iterative pursuit reintroducing precisely the computational cost that
\ours{} eliminates.

\subsection{Spectral Initialization}
\label{sec:spectral}

We compute the top-$k$ eigenvectors of $\m{W}_L\m{W}_L^{\top}$:
\begin{equation}
  \m{W}_L\m{W}_L^{\top}\approx
  \m{B}\boldsymbol{\Lambda}_k\m{B}^{\top},
  \qquad
  \m{B}\in\R^{d_1\times k},\;
  \m{B}^{\top}\m{B}=\m{I}_k,
  \qquad
  \m{C}:=\m{B}^{\top}\m{W}_L\in\R^{k\times d_2},
  \label{eq:evd}
\end{equation}
so that $\m{W}_L\approx\m{B}\m{C}$.  By the
Eckart--Young--Mirsky theorem, $\m{B}$ is the optimal
column-orthogonal basis of rank~$k$: no rank-$k$ approximation
achieves a smaller whitened reconstruction error, and the associated
coefficients $\m{C}$ come for free as orthogonal projections, with no pursuit required.

Because $\m{B}$ is semi-orthogonal, zeroing $c_{ij}$ deactivates the $i$-th latent direction for the $j$-th output dimension.  Since sparsity is applied independently per column, each output may activate a \emph{distinct} subset of basis vectors, realizing a union-of-subspaces model similar to dictionary learning, while retaining
a single shared basis $\m{B}$ that is reused across all candidate configurations (Section~\ref{sec:profiling}).

\subsection{Support Selection by Calibrated Hard Thresholding}
\label{sec:thresholding}

\paragraph{Why plain thresholding is not enough.}
For an orthonormal basis $\m{B}$, the whitened reconstruction error satisfies the Pythagorean decomposition $\norm{\m{W}_L-\m{B}\m{S}}{F}^{2}
=\norm{\m{W}_L-\m{B}\m{C}}{F}^{2}+\norm{\m{C}-\m{S}}{F}^{2}$, so hard thresholding of $|c_{ij}|$ is the exact best-$s$-term solver~\cite{mallat1999wavelet,elad2010sparse}.
However, the whitening transform is estimated from a finite calibration set, so it captures only the observed statistics of the input activations and cannot fully cover the input distribution. To avoid relying exclusively on this calibration-dependent whitened geometry, we balance reconstruction fidelity in the whitened space with reconstruction quality in the original weight space. Since $\m{L}^{-1}$ is generally non-orthogonal, two coefficients with identical whitened magnitudes may contribute very differently to the reconstructed weight matrix. Specifically, an error of unit whitened norm along $\m{b}_i$ incurs an original-space cost 
$a_i:=\norm{\m{L}^{-1}\m{b}_i}{2}$.

\paragraph{Fused importance score.}
To correct for this metric mismatch we define two complementary importance measures:
\begin{equation}
  \mathrm{imp}^{\mathrm{white}}_{ij}=|c_{ij}|,
  \qquad
  \mathrm{imp}^{\mathrm{orig}}_{ij}
  =|c_{ij}|\cdot\norm{\m{L}^{-1}\m{b}_i}{2},
  \label{eq:imp_orig}
\end{equation}
and fuse them via a scale-invariant geometric interpolation with balance parameter $\lambda\in[0,1]$:
\begin{equation}
  \mathrm{imp}_{ij}
  =\bigl(\mathrm{imp}^{\mathrm{white}}_{ij}\bigr)^{1-\lambda}
   \bigl(\mathrm{imp}^{\mathrm{orig}}_{ij}\bigr)^{\lambda}
  =|c_{ij}|\cdot\norm{\m{L}^{-1}\m{b}_i}{2}^{\lambda}.
  \label{eq:imp_final}
\end{equation}
Let $\boldsymbol{\nu}\in\R^{k}$ with $\nu_i=\norm{\m{L}^{-1}\m{b}_i}{2}^{\lambda}$; the full importance matrix is
\begin{equation}
  \mathbf{Imp}=|\m{C}|\odot
  \bigl(\boldsymbol{\nu}\,\mathbf{1}_{d_2}^{\top}\bigr).
  \label{eq:imp_matrix}
\end{equation}
This scoring is the solution of a weighted sparse approximation problem in which the weights encode the distortion induced by $\m{L}^{-1}$; the derivation, the exact quadratic form, and its diagonal surrogate
are given in Appendix~\ref{app:weighted-sap}.  The surrogate neglects off-diagonal couplings in
$\m{G}=\m{B}^{\top}\m{A}^{-1}\m{B}$; these couplings vanish when the
un-whitened atoms $\{\m{L}^{-1}\m{b}_i\}$ are mutually orthogonal, and empirically the trade-off is favourable
(Section~\ref{sec:ablations}).

\paragraph{Two-stage thresholding.}
We apply column-wise hard thresholding on $\m{C}$, retaining the top-$s$ entries per column according to $\mathrm{imp}_{ij}$.  To allow global flexibility beyond fixed per-column budgets, we initially over-sparsify $\m{C}$ by a small percentage $\gamma$
($\gamma=5\times10^{-3}$ in all experiments), then reactivate the most important masked coefficients across the entire matrix until the exact target compression ratio $\mathrm{cr}$ is reached.  For every fixed $\lambda$, thresholding by $\mathrm{imp}_{ij}$ is the optimizer of the
stated weighted objective; the selection step never requires a greedy pursuit, in contrast to OMP-based coding for general dictionaries. This yields the sparse coefficient matrix $\m{C}_{\mathrm{sparse}}$.
We set $\lambda=0.5$ as default; the full sensitivity analysis over $\lambda\in[0,1]$ is provided in
Section~\ref{sec:lambda_ablation}.

\subsection{ Dictionary Refinement}
\label{sec:refinement}

Orthogonality of $\m{B}$ was needed only to make the thresholding step closed-form.  Once the support of $\m{C}_{\mathrm{sparse}}$ is fixed,
the dictionary can be fit freely.  We compute the final dictionary $\m{D}_{\mathrm{final}}$ by solving a ridge-regularized least-squares problem:
\begin{equation}
  \m{D}_{\mathrm{final}}
  =\argmin_{\m{D}}\;
  \norm{\m{W}_L-\m{D}\m{C}_{\mathrm{sparse}}}{F}^{2}
  +\mu\norm{\m{D}}{F}^{2},
  \label{eq:ls}
\end{equation}
which admits a closed-form Cholesky-based solution; the adaptive ridge
$\mu$ is specified in Appendix~\ref{hyp_ablation_details}.  This is the
MOD dictionary update~\cite{engan1999method} computed for the fixed
code $\m{C}_{\mathrm{sparse}}$: whereas K-SVD updates atoms
sequentially, \ours{} evaluates the joint minimizer given the codes,
yielding a better fit without increasing inference cost.  The final
compressed weight is recovered as
\begin{equation}
  \widetilde{\m{W}}
  =\m{L}^{-1}\m{D}_{\mathrm{final}}\m{C}_{\mathrm{sparse}},
  \qquad\text{stored as}\;\;
  \m{F}=\m{L}^{-1}\m{D}_{\mathrm{final}},\;\;
  \m{H}=\m{C}_{\mathrm{sparse}}.
  \label{eq:recover}
\end{equation}
When no sparsity is applied ($s=k$), thresholding reduces to the identity operator
and \ours{} recovers standard data-aware low-rank SVD
(Appendix~\ref{app:svd-recovery}).

Because each step is closed-form and the basis is computed once per
layer, the entire factorization of a 448-layer model completes in
under 2~minutes on a single CPU thread
(Appendix~\ref{sec:scalability_runtime}), compared to 646~min for
iterative K-SVD/OMP (CoSpaDi).
\paragraph{Compression-ratio convention and parameter counting.}
Throughout the paper, the compression ratio ($\mathrm{CR}$) denotes the fraction of parameters \emph{removed}. Thus, $\mathrm{CR}=0.2$ means that $20\%$ of the targeted compressible parameters are removed and $80\%$ are retained. For a linear layer $\m W\in\mathbb{R}^{d_1\times d_2}$ compressed with dictionary size $k$ and column sparsity $s$, where $s$ is the number of nonzeros retained per column of $\m C_{\mathrm{sparse}}$, ROCKET stores the dense factor $\m F\in\mathbb{R}^{d_1\times k}$ and the sparse coefficient matrix $\m H=\m C_{\mathrm{sparse}}\in\mathbb{R}^{k\times d_2}$. The number of stored numerical parameters is therefore
$
P_{\mathrm{stored}} = d_1 k + s d_2,
$
and the corresponding layer-wise removed fraction is
$
\mathrm{CR}_{\ell} = 1 - \frac{d_1 k + s d_2}{d_1 d_2}.
$
The retained fraction is $1-\mathrm{CR}$. Sparse indices for $\m H$ are accounted for by truncating the 16-bit floating-point representation by two bits, following CoSpaDi~\cite{Shopkhoev2025cospadi0}. For a global target compression ratio $\mathrm{CR}$, the global kept-parameter budget is $(1-\mathrm{CR})P_{\mathrm{orig}}$, where $P_{\mathrm{orig}}$ is the total number of parameters in the compressible linear layers.
% To enable optimal global compression under a fixed parameter budget, we first perform a lightweight \textit{layer profiling} pass. For each compressible layer, we evaluate a predefined set of (rank, sparsity) configurations. For each candidate, we: (i) compute the eigendecomposition of $W_LW_L^\top$, (ii) determine rank $k$ and sparsity level $s$, (iii) sparsify $C$ using fused importance scores, (iv) compute $D_{\mathrm{final}}$ via least squares, and (v) record the stored parameter count $c_{\ell,i}=d_{1,\ell}k_{\ell,i}+s_{\ell,i}d_{2,\ell}$, the $k/s$ ratio, and the relative reconstruction error ($e_{\ell,i}\leq 1$) in the original space for layer $\ell$ and option $i$, given as $e_{\ell,i}=\|W_{\ell}-\widetilde{W}_{\ell i}\|_F/\|W_{\ell}\|_F$. Here $c_{\ell,i}$ is the kept parameter count for option $i$ of layer $\ell$, and the corresponding layer-wise removed fraction is $1-c_{\ell,i}/(d_{1,\ell}d_{2,\ell})$.
% This yields, per layer $\ell$, a discrete set of feasible options $\mc{O}_\ell = \{(c_{\ell,i}, ks_{\ell,i}, e_{\ell,i})\}$.
% During the profiling stage, the eigenvalue decomposition (EVD) is computed exactly once per layer. This allows reconstruction errors across varying compression configurations to be evaluated efficiently without recomputing the decomposition. Specifically, error estimation is achieved by traversing the precomputed spectral components and applying a hard-thresholding operation to the coefficient matrix. For a detailed runtime analysis, we refer to Appendix 
\paragraph{Layer Profiling.}
\label{sec:profiling}
To select the best compression configuration under a global parameter budget, ROCKET first performs a lightweight \textit{layer-profiling} pass over all compressible layers. For each layer $\ell$, we consider a predefined set of candidate configurations, where each candidate $i$ is specified by a rank $k_{\ell,i}$ and a coefficient-sparsity level $s_{\ell,i}$. For each candidate, we: (i) compute the eigendecomposition of $\m W_{L,\ell}\m W_{L,\ell}^\top$; (ii) retain the top $k_{\ell,i}$ basis vectors; (iii) form the coefficient matrix $\m C_{\ell,i}$ and sparsify it using the fused importance scores; (iv) compute the refined dictionary $\m D_{\mathrm{final},\ell,i}$ via closed-form least squares; and (v) record the stored parameter count $
c_{\ell,i}=d_{1,\ell}k_{\ell,i}+s_{\ell,i}d_{2,\ell},
$
the rank-to-sparsity ratio $ks_{\ell,i}=k_{\ell,i}/s_{\ell,i}$, and the relative reconstruction error in the original space,
$
e_{\ell,i}=\frac{\|\m W_{\ell}- \widetilde{\m W}_{\ell,i}\|_F}{\|\m W_{\ell}\|_F}\leq 1.
$
Here, $d_{1,\ell}$ and $d_{2,\ell}$ denote the input and output dimensions of layer $\ell$, and $c_{\ell,i}$ is the number of parameters kept by candidate $i$. The corresponding layer-wise compression ratio is therefore
$
\mathrm{CR}_{\ell,i}=1-\frac{c_{\ell,i}}{d_{1,\ell}d_{2,\ell}},
$
where $\mathrm{CR}$ denotes the fraction of parameters removed. This procedure yields, for each linear layer $\ell$, a discrete set of feasible compression options
$
\mathcal{O}_\ell=\{(c_{\ell,i},ks_{\ell,i},e_{\ell,i})\}_{i=1}^{K_\ell}.
$
Importantly, the eigendecomposition is computed only once per layer. All subsequent candidates are evaluated by reusing the precomputed spectral components and applying hard thresholding to the coefficient matrix, which avoids recomputing the decomposition and makes profiling efficient. A detailed runtime analysis is provided in Appendix \ref{sec:scalability_runtime}; the derivation of the reconstruction error metric is provided in Appendix \ref{reconstruction_error_dis}.
The final output of the profiling stage is the collection $\{\mathcal{O}_\ell\}_{\ell=1}^{L}$, where each layer $\ell$ is associated with a discrete menu of feasible compression options. Each option records its stored-parameter cost $c_{\ell,i}$, rank-to-sparsity ratio $ks_{\ell,i}$, and reconstruction error $e_{\ell,i}$. These candidate sets are then passed to the constrained multi-choice knapsack solver, which selects exactly one option per layer under the global compression budget.

\paragraph{Constrained Multi-Choice Knapsack Formulation.}
\label{sec:dp}
Let there be $L$ compressible layers. For each layer $\ell \in \{1, \dots, L\}$, the profiling stage produces a discrete set of feasible compression options
$
\mathcal{O}_\ell = \{(c_{\ell,i},\, ks_{\ell,i},\, e_{\ell,i})\}_{i=1}^{K_\ell},
$
Let $P_{\mathrm{orig}}$ be the total number of parameters in the compressible layers.
For a target compression ratio $\mathrm{CR}$, the global kept-parameter budget is
$
C_{\mathrm{total}} = (1 - \mathrm{CR})\,P_{\mathrm{orig}}.
$
The task of selecting exactly one option per layer so as to minimize total reconstruction error while respecting this budget is a \textit{multi-choice knapsack problem} (MCKP); an ablation on the discretization step is provided in Section~\ref{subsec:ablation_search}:
\bal
&\min_{x_{\ell,i} \in \{0,1\}} \sum_{\ell=1}^L \sum_{i=1}^{K_\ell} e_{\ell,i}\, x_{\ell,i}
\quad \text{s.t.} \quad
\sum_{\ell=1}^L \sum_{i=1}^{K_\ell} c_{\ell,i}\, x_{\ell,i} \leq C_{\mathrm{total}}, \quad
\sum_{i=1}^{K_\ell} x_{\ell,i} = 1,\; \forall\, \ell.\nonumber
\eal

To prevent pathological allocations that achieve low global error by severely damaging a few layers while over-preserving others, we impose an additional per-layer error cap.
Let $\bar{e}_{\mathrm{ref}}$ denote the average reconstruction error obtained when every layer is compressed at a fixed reference ratio $\rho_{\mathrm{ref}}$.
We then require that the error of the selected option for each layer does not exceed a scaled version of this reference:
$
\sum_{i=1}^{K_\ell} e_{\ell,i}\, x_{\ell,i} \leq \alpha\,\bar{e}_{\mathrm{ref}}
\quad \forall\, \ell \in \{1, \dots, L\},
$
where $\alpha$ is a tunable hyperparameter.
In all experiments, we set $\alpha = \alpha_{\min} := \inf\{\alpha' > 0 \mid \text{a feasible solution exists for } \alpha'\}$,
i.e., the smallest cap that still admits a valid allocation.
Further per-model tuning of $\alpha$ may yield additional improvements.
With these layer-wise error caps, the problem remains an MCKP whose solution is globally optimal over the discretized candidate set and ensures local robustness of each individual layer.
We reformulate this problem as a shortest-path search on a directed acyclic graph in Appendix~\ref{graphtheorymckp}, which clarifies the dynamic-programming solution that follows.\vspace{-.2cm}
\section{Experiments} 
\vspace{-0.0cm}
\begin{table*}[h]
\caption{Performance comparison of \ours{} vs SOTA SVD-LLM and CoSpaDi methods on Qwen3-8B at different compression ratios (CR). Best results are highlighted with \textbf{bold}.}
\label{tab:main_qwen3}
\centering
\resizebox{0.95\textwidth}{!}{%
\renewcommand{\arraystretch}{1.1}
\begin{tabular}{ccccccccccccc}
\hline
                            &                          & \multicolumn{9}{c}{\textbf{Accuracy$\uparrow$}}                                                            & \multicolumn{2}{c}{\textbf{Perplexity$\downarrow$}}       \\ \cline{3-13}
\multirow{-2}{*}{\textbf{Method}} & \multirow{-2}{*}{\textbf{CR}} & \textbf{PIQA} & \textbf{HellaSwag} & \textbf{LAMBADA} & \textbf{ARC-e} & \textbf{ARC-c} & \textbf{SciQ} & \textbf{Race} & \textbf{MMLU} & \textbf{Avg.} & \textbf{WikiText} & \textbf{LAMBADA} \\ \hline

\textbf{Qwen3 8B} & -- & 77.7 & 74.9 & 64.1 & 80.7 & 56.7 & 95.7 & 40.9 & 73.0 & 70.5 & 12 & 4.6 \\ \hline

SVD-LLM &  & 73.8 & 63.9 & 62.2 & 68.7 & 45.7 & 90.1 & 40.5 & 54.7 & 62.5 & 21 & 6.4 \\
CoSpaDi &  & 76.5 & 68.0 & 65.6 & 72.2 & 48.9 & 93.2 & 40.7 & 60.8 & 65.7 & 18 & 4.9 \\
ROCKET  & \multirow{-3}{*}{0.2}
        & \textbf{77.6} & \textbf{72.9} & \textbf{66.0} & \textbf{75.8} & \textbf{53.9} & \textbf{94.5} & \textbf{41.4} & \textbf{67.2} & \textbf{68.7} & \textbf{15} & \textbf{4.7} \\ \hline

SVD-LLM &  & 70.4 & 55.2 & 53.8 & 59.3 & 37.1 & 87.2 & 38.4 & 44.8 & 55.8 & 27 & 11 \\
CoSpaDi &  & 72.4 & 60.5 & 62.6 & 63.9 & 41.2 & 88.4 & 39.5 & 51.3 & 60.0 & 23 & 6.3 \\
ROCKET  & \multirow{-3}{*}{0.3}
        & \textbf{75.1} & \textbf{67.2} & \textbf{68.0} & \textbf{72.1} & \textbf{47.7} & \textbf{92.9} & \textbf{41.4} & \textbf{62.3} & \textbf{65.8} & \textbf{18} & \textbf{4.4} \\ \hline

SVD-LLM &  & 66.3 & 44.6 & 37.9 & 45.0 & 28.1 & 77.3 & 35.3 & 29.1 & 45.4 & 43 & 36 \\
CoSpaDi &  & 68.9 & 49.0 & 49.9 & 49.4 & 29.9 & 82.0 & 36.8 & 36.6 & 50.3 & 36 & 15 \\
ROCKET  & \multirow{-3}{*}{0.4}
        & \textbf{71.0} & \textbf{58.7} & \textbf{63.7} & \textbf{65.7} & \textbf{40.4} & \textbf{86.1} & \textbf{41.9} & \textbf{52.3} & \textbf{60.0} & \textbf{24} & \textbf{5.9} \\ \hline

SVD-LLM &  & 61.5 & 34.9 & 25.1 & 37.4 & 25.3 & 65.1 & 31.6 & 24.0 & 38.1 & 76 & 88 \\
CoSpaDi &  & 63.8 & 39.7 & 32.4 & 41.2 & 26.8 & 70.4 & 33.2 & 28.1 & 42.0 & 59 & 41 \\
ROCKET  & \multirow{-3}{*}{0.5}
        & \textbf{68.8} & \textbf{48.4} & \textbf{47.5} & \textbf{54.8} & \textbf{33.0} & \textbf{81.6} & \textbf{38.5} & \textbf{37.9} & \textbf{51.3} & \textbf{35} & \textbf{24} \\ \hline

\end{tabular}
}
\end{table*}
\vspace{-0.3 cm}
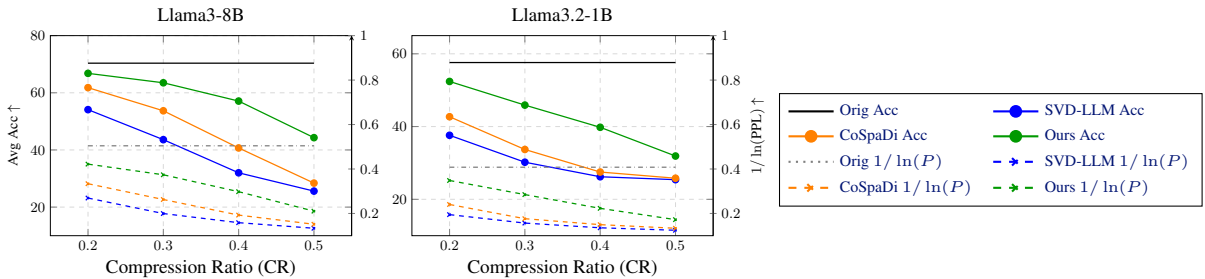
\begin{figure*}[b]
  \centering
  \small % Makes text slightly smaller to fit better side-by-side
\resizebox{0.65\textwidth}{!}{
  % --- LLAMA3-8B (LEFT) ---
  \begin{tikzpicture}
    % \begin{axis}[
    %   width=0.48\textwidth, height=6cm,
    %   title={Llama3-8B},
    %   xlabel={Compression Ratio (CR)},
    %   ylabel={Avg Acc $\uparrow$},
    %   ymin=10, ymax=80, xmin=0.15, xmax=0.55,
    %   grid=major, grid style={dashed, gray!30},
    %   % Export the legend to a global name
    %   legend to name=sharedlegend,
    %   legend columns=4, 
    %   legend style={font=\tiny, draw=black, fill=white, /tikz/every even column/.append style={column sep=0.2cm}},
    %   legend cell align={left},
    % ]
    %   \addplot[black, solid, thick, domain=0.2:0.5] {70.36}; \addlegendentry{Orig Acc}
    %   \addplot[blue, mark=*, thick] coordinates {(0.2,54.1) (0.3,43.6) (0.4,32.0) (0.5,25.6)}; \addlegendentry{SVD Acc}
    %   \addplot[orange, mark=*, thick] coordinates {(0.2,61.8) (0.3,53.7) (0.4,40.7) (0.5,28.4)}; \addlegendentry{CoSpa Acc}
    %   \addplot[green!60!black, mark=*, thick] coordinates {(0.2,66.8) (0.3,63.5) (0.4,57.1) (0.5,44.3)}; \addlegendentry{Ours Acc}
      
    %   % Placeholders for Legend
    %   \addlegendimage{gray, dotted, thick} \addlegendentry{Orig $1/\ln(P)$}
    %   \addlegendimage{blue, dashed, mark=x, thick} \addlegendentry{SVD $1/\ln(P)$}
    %   \addlegendimage{orange, dashed, mark=x, thick} \addlegendentry{CoSpa $1/\ln(P)$}
    %   \addlegendimage{green!60!black, dashed, mark=x, thick} \addlegendentry{Ours $1/\ln(P)$}
    % \end{axis}
    \begin{axis}[
      width=0.48\textwidth, height=6cm,
      title={Llama3-8B},
      title style={font=\large},
      xlabel={Compression Ratio (CR)},
      xlabel style={font=\large},
      ylabel={Avg Acc $\uparrow$},
      ymin=10, ymax=80, xmin=0.15, xmax=0.55,
      grid=major, grid style={dashed, gray!30},
      % Export the legend to a global name
      legend to name=sharedlegend,
      legend columns=1,  % Изменено с 4 на 2 → 4 строки × 2 столбца
      legend style={font=\tiny, draw=black, fill=white, /tikz/every even column/.append style={column sep=0.2cm}},
      legend cell align={left},
    ]
      \addplot[black, solid, thick, domain=0.2:0.5] {70.36}; \addlegendentry{Orig Acc}
      \addplot[blue, mark=*, thick] coordinates {(0.2,54.1) (0.3,43.6) (0.4,32.0) (0.5,25.6)}; \addlegendentry{SVD-LLM Acc}
      \addplot[orange, mark=*, thick] coordinates {(0.2,61.8) (0.3,53.7) (0.4,40.7) (0.5,28.4)}; \addlegendentry{CoSpaDi Acc}
      \addplot[green!60!black, mark=*, thick] coordinates {(0.2,66.8) (0.3,63.5) (0.4,57.1) (0.5,44.3)}; \addlegendentry{Ours Acc}
      
      % Placeholders for Legend
      \addlegendimage{gray, dotted, thick} \addlegendentry{Orig $1/\ln(P)$}
      \addlegendimage{blue, dashed, mark=x, thick} \addlegendentry{SVD-LLM $1/\ln(P)$}
      \addlegendimage{orange, dashed, mark=x, thick} \addlegendentry{CoSpaDi $1/\ln(P)$}
      \addlegendimage{green!60!black, dashed, mark=x, thick} \addlegendentry{Ours $1/\ln(P)$}
    \end{axis}
    \begin{axis}[
      width=0.48\textwidth, height=6cm,
      axis y line=right, axis x line=none,
      ymin=0.1, ymax=1.0, xmin=0.15, xmax=0.55,
    ]
      \addplot[gray, dashdotted, thick, domain=0.2:0.5] {0.5044};
      \addplot[blue, dashed, mark=x, thick] coordinates {(0.2,0.2693) (0.3,0.1996) (0.4,0.1585) (0.5,0.1334)};
      \addplot[orange, dashed, mark=x, thick] coordinates {(0.2,0.3338) (0.3,0.2627) (0.4,0.1926) (0.5,0.1514)};
      \addplot[green!60!black, dashed, mark=x, thick] coordinates {(0.2,0.4219) (0.3,0.3740) (0.4,0.2979) (0.5,0.2103)};
    \end{axis}
  \end{tikzpicture}
  \hfill % Pushes the next graph to the right
  % --- LLAMA3.2-1B (RIGHT) ---
  \begin{tikzpicture}
    \begin{axis}[
      width=0.48\textwidth, height=6cm,
      title={Llama3.2-1B},
      title style={font=\large},
      xlabel={Compression Ratio (CR)},
      xlabel style={font=\large},
      ylabel style={align=center}, 
      ymin=10, ymax=65, xmin=0.15, xmax=0.55,
      grid=major, grid style={dashed, gray!30},
      % No legend defined here as it is shared
    ]
      \addplot[black, solid, thick, domain=0.2:0.5] {57.6}; 
      \addplot[blue, mark=*, thick] coordinates {(0.2,37.6) (0.3,30.2) (0.4,26.2) (0.5,25.4)}; 
      \addplot[orange, mark=*, thick] coordinates {(0.2,42.7) (0.3,33.7) (0.4,27.5) (0.5,25.8)}; 
      \addplot[green!60!black, mark=*, thick] coordinates {(0.2,52.4) (0.3,45.9) (0.4,39.8) (0.5,31.9)}; 
    \end{axis}
    \begin{axis}[
      width=0.48\textwidth, height=6cm,
      axis y line=right, axis x line=none,
      ylabel={$1 / \ln(\text{PPL}) \uparrow$},
      ymin=0.1, ymax=1.0, xmin=0.15, xmax=0.55,
    ]
      \addplot[gray, dashdotted, thick, domain=0.2:0.5] {0.4080};
      \addplot[blue, dashed, mark=x, thick] coordinates {(0.2,0.1947) (0.3,0.1567) (0.4,0.1355) (0.5,0.1244)};
      \addplot[orange, dashed, mark=x, thick] coordinates {(0.2,0.2404) (0.3,0.1764) (0.4,0.1496) (0.5,0.1334)};
      \addplot[green!60!black, dashed, mark=x, thick] coordinates {(0.2,0.3483) (0.3,0.2849) (0.4,0.223) (0.5,0.1722)};
    \end{axis}
  \end{tikzpicture}
}
  \raisebox{0.3cm}{\ref{sharedlegend}} % This prints the legend defined in the first plot
  
  \caption{Comparison of Accuracy and Inverse Log Perplexity for Llama3-8B and Llama3.2-1B.}
  \label{othmodels}\vspace{-.7cm}
\end{figure*}

In this section, we describe our experimental setup and compare \textbf{\ours{}} against recent  compression methods. We focus on low-rank and dictionary-based approaches, specifically SVD-LLM~\cite{wang2025svdllm} and CoSpaDi~\cite{Shopkhoev2025cospadi0}, as well as budget allocation based methods including ARS~\cite{ars}, Dobi-SVD~\cite{dobisvd}, and ARA~\cite{ara}. We also provide comparisons with sparsification and width-pruning methods such as LLM-Pruner~\cite{llmpruner}, SliceGPT~\cite{slicegpt}, Bonsai~\cite{bonsai}, and Wanda~\cite{wanda}. Finally, we conduct ablation studies to isolate the contribution of each one of our design choices.\vspace{-.3cm}

\subsection{Experimental Setup}\vspace{-0.1cm}
\label{sec:exp_setup}
We evaluate our method in a per-layer setting using Llama and Qwen models. All evaluations are performed in a zero-shot setting on the following benchmarks: PIQA~\cite{bisk2019piqareasoningphysicalcommonsense}, HellaSwag~\cite{zellers2019hellaswagmachinereallyfinish}, OpenAI LAMBADA~\cite{paperno2016lambadadatasetwordprediction}, ARC-Easy and ARC-Challenge~\cite{clark2018think}, SciQ~\cite{welbl2017crowdsourcingmultiplechoicescience}, RACE~\cite{lai2017racelargescalereadingcomprehension}, and MMLU~\cite{hendrycks2021measuringmassivemultitasklanguage}. In addition, we report perplexity on WikiText~\cite{wikitext} and LAMBADA-OpenAI.
We apply compression at compression ratios ranging from 0.2 to 0.5, in steps of 0.1. For methods that need calibration data, we use 256 randomly sampled sequences from the FineWeb dataset~\cite{refinedweb} (fixed across all experiments). We also test how the choice of calibration dataset affects results in Appendix~\ref{data_ablation}. Unless otherwise noted, we compress all dense linear layers in the self-attention blocks (Q, K, V, and O projections) and the feed-forward network (gate, up, and down projections). Embedding layers and the \texttt{lm\_head} are not compressed following other works. All baselines are reproduced with official code under the same setup, except CoSpaDi in Table~\ref{tab:main_qwen3}, taken from~\cite{Shopkhoev2025cospadi0}, and the baselines in Table~\ref{tab:llama3_results}, taken from the respective papers. More experiments are reported in Appendix~\ref{more_results}.\vspace{-.3cm}

\paragraph{Comparison with SVD-LLM and CoSpaDi}

To contextualize ROCKET’s performance, we directly compare it against SVD-LLM and CoSpaDi, the two most closely related training-free compression methods. All methods are evaluated in a strictly training-free setting, no fine-tuning, healing, or data augmentation is applied post-compression.
As shown in Table~\ref{tab:main_qwen3} and Fig.~\ref{othmodels}, \ours{} consistently outperforms both baselines by a significant margin across multiple architectures (Qwen3-8B, Llama3-8B and Llama3.2-1B) and compression ratios (20\%–50\%), in terms of both zero-shot accuracy and perplexity (we refer to Appendix \ref{detailed_comparison} for more detailed results). Notably, \ours{} exhibits superior scalability under aggressive compression and increasing model scales (details are provided in Appendix \ref{scaling_property}): while baseline methods suffer severe degradation beyond 30\% compression, \ours{} retains more robust performance even at 50\% compression (e.g., 51.3 average accuracy on Qwen3-8B vs. 38.1 for SVD-LLM and 42.0 for CoSpaDi). This demonstrates that \ours{}’s combination of calibration-guided factorization, importance-based sparsification and knapsack-based layer-wise budget allocation effectively preserves model fidelity under strict parameter constraints.\vspace{-.2cm}

\paragraph{Comparison against other budget allocation methods} 

\begin{figure*}[h]
    \centering
    \includegraphics[width=0.6\linewidth]{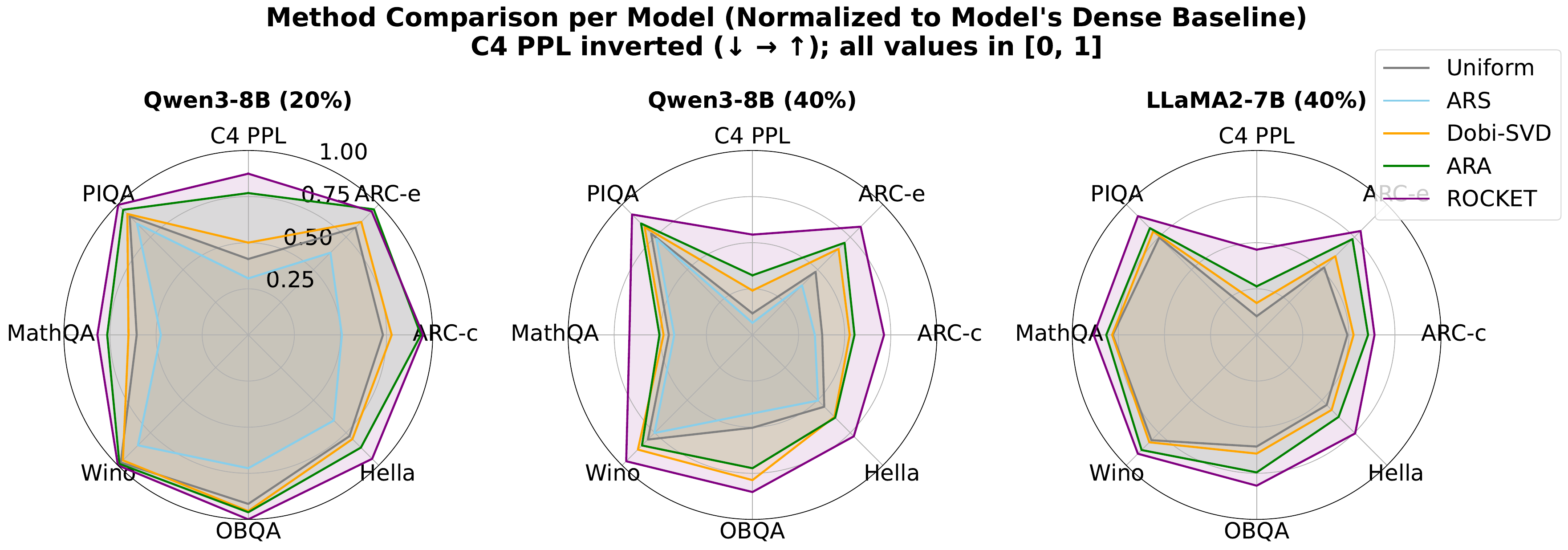}
    \caption{Comparison of \ours{} with alternative budget allocation methods (Uniform, ARS~\cite{ars}, Dobi-SVD~\cite{dobisvd}, and ARA~\cite{ara}) on three model configurations: Qwen3-8B at 20\% and 40\% pruning, and Llama2-7B at 40\% pruning. Subplots show normalized performance on eight benchmarks (C4 perplexity is inverted, meaning higher value is better), scaled to each model’s dense baseline (value=1.0). \ours{} consistently retains the best performance.}
    \label{against_budget_aloc}\vspace{-.3cm}
\end{figure*}
To evaluate the effectiveness of ROCKET’s layer-wise budget allocation, we compare it against four established parameter allocation strategies: SVD-LLM~\cite{wang2025svdllm}, which applies uniform compression across layers; Adaptive Rank Selection (ARS)~\cite{ars}; Dobi-SVD~\cite{dobisvd}; and Adaptive Rank Allocation (ARA)~\cite{ara}. Figure~\ref{against_budget_aloc} shows normalized performance across three model-compressions configurations Qwen3-8B at 20\% and 40\% compression, and Llama2-7B at 40\% compression with all scores scaled to their respective dense baselines. ROCKET consistently outperforms all baselines, demonstrating that globally optimizing the allocation of parameters via constrained knapsack selection preserves significantly more of the original model’s capabilities than uniform or trainable strategies, especially under aggressive compression and across different architectures.
To further contextualize ROCKET's knapsack-based allocation, we compare it against HALP~\cite{shen2021halp}, a hardware-aware latency pruning method, applied on top of dictionary learning (denoted DictionaryLearning+HALP). On Llama3.2-1B at a 20\% compression ratio, DictionaryLearning+HALP obtains 30.38 average accuracy and 285 perplexity, whereas ROCKET achieves 52.4 average accuracy and 18 perplexity. This demonstrates that ROCKET's training-free multi-choice reconstruction-error allocation  outperforms HALP-style structured pruning under the same compression budget.

\paragraph{Comparison with Depth and Sparsity Pruning Methods}
\label{depth_sparsity_pruning}

To demonstrate the effectiveness of our method, we compare it against several approaches that address model compression from different perspectives. These include width pruning (e.g., SliceGPT~\cite{slicegpt}), combined depth/width pruning (e.g., LLM-Pruner~\cite{llmpruner}), structured sparsification (e.g., Wanda~\cite{wanda}, Bonsai), adaptive low-rank decomposition with quantization (e.g., Dobi-SVD~\cite{dobisvd}), and training-free modular decomposition (e.g., MoDeGPT~\cite{lin2025modegpt}). As shown in Table~\ref{tab:llama3_results}, ROCKET achieves strong performance, outperforming all other non-quantized baselines at a 40\% compression ratio (0.56 average accuracy). Since Dobi-SVD uses quantization, we additionally evaluate ROCKET with post-compression quantization to ensure a fair comparison (weight-only INT8 min-max quantization of both factors without calibration; Appendix~\ref{app:quantization}). Under this setting, ROCKET surpasses Dobi-SVD at a 40\% compression ratio (0.65 vs.\ 0.63). At a 60\% compression ratio, ROCKET again leads with an average accuracy of 0.60 compared to Dobi-SVD's 0.52. To further compare with a training-free structured decomposition baseline, we evaluate MoDeGPT on Llama3-8B at a 25\% compression ratio under the same training-free protocol. The dense Llama3-8B model achieves 72.75\% average accuracy, while MoDeGPT retains approximately 88\% of this performance with 63.96\% average accuracy. In contrast, ROCKET retains approximately 95\%, achieving 69.34\% average accuracy. This demonstrates that ROCKET's training-free pipeline is effective, matching or exceeding other pruning and structured decomposition methods.

\begin{table*}[h]
\centering
\caption{Performance comparison of \ours{} against depth- and sparsity-based pruning methods on Llama3.1-8B across compression levels and benchmarks. ``Avg'' denotes the average accuracy across all benchmarks, ``Drop'' indicates the relative accuracy drop percentage compared to the dense model, and ``Quant.'' indicates whether post-compression quantization is applied to achieve the target CR.}
\label{tab:llama3_results}
\resizebox{0.9\textwidth}{!}{%
\begin{tabular}{l|c|l|c|c|c|c|c|c|c|c}
\toprule
\textbf{Models} & \textbf{CR} & \textbf{Method} & \textbf{Quant.} & \textbf{PIQA} & \textbf{HellaSwag} & \textbf{WinoGrande} & \textbf{ARC\_e} & \textbf{ARC\_c} & \textbf{\begin{tabular}[c]{@{}c@{}}Avg. ($\uparrow$)\end{tabular}} & \textbf{\begin{tabular}[c]{@{}c@{}}Drop ($\downarrow$)\end{tabular}} \\
\midrule
\multirow{10}{*}{\textbf{Llama3.1-8B}} 
& -- & Baseline & \ding{55} & 0.80 & 0.59 & 0.74 & 0.81 & 0.51 & 0.69 & 0\% \\
\cmidrule{2-11}
& \multirow{5}{*}{0.4} & LLM-Pruner & \ding{55} & 0.66 & 0.32 & 0.54 & 0.58 & 0.23 & 0.46 & 33.3\% \\
& & SliceGPT & \ding{55} & 0.62 & 0.40 & 0.53 & 0.49 & 0.25 & 0.46 & 33.3\% \\
& & Bonsai & \ding{55} & 0.59 & 0.29 & 0.49 & 0.47 & 0.18 & 0.41 & 40.6\% \\
& & Wanda-sp & \ding{55} & 0.57 & 0.28 & 0.50 & 0.44 & 0.17 & 0.39 & 43.5\% \\
& & Dobi-SVD & \checkmark & 0.76 & 0.52 & 0.72 & 0.73 & 0.39 & 0.63 & 8.70\% \\
& & \ours & \ding{55} & 0.72 & 0.43 & 0.66 & 0.64 & 0.33 & 0.56 & 18.84\% \\
& & \ours & \checkmark & \textbf{0.78} & \textbf{0.54} & \textbf{0.72} & \textbf{0.76} & \textbf{0.42} & \textbf{0.65} & \textbf{5.79}\% \\
\cmidrule{2-11}
& 0.6 & Dobi-SVD & \checkmark & 0.68 & 0.41 & 0.66 & 0.58 & 0.27 & 0.52 & 24.6\% \\
&  & \ours & \checkmark & \textbf{0.75} & \textbf{0.49} & \textbf{0.68} & \textbf{0.72} & \textbf{0.38} & \textbf{0.60} & \textbf{13.0\%} \\
\bottomrule
\end{tabular}
}
\end{table*}

\paragraph{Post-Compression Healing}
Applying a lightweight healing step that fine-tunes only unmasked parameters on 30M C4 tokens while preserving the fixed sparsity pattern recovers substantial fidelity in the 40\%-compressed ROCKET-Qwen3-14B model, increasing average accuracy from 63.56 to 67.96 and reducing perplexity from $25$ to $13$. This healed configuration closely approaches the native Qwen3-8B baseline (70.46 accuracy, $12$ perplexity), confirming that ROCKET provides a robust initialization for efficient post-compression recovery. We refer to Appendix \ref{healing_detailed} for more details.
\paragraph{Generalization to Other Modalities}
To assess the generality of ROCKET beyond language-only models, we apply it to two transformer-based architectures from different modalities: the vision-language model \textbf{Qwen3-VL-4B} and the speech generation model \textbf{VibeVoice} \cite{vibevoice}. For VibeVoice, we use 256 transcriptions-only from \texttt{mls\_eng\_10k} dataset and validate speech results on different samples from \texttt{mls\_eng\_10k}~\cite{tts_benchmark}. For Qwen3-VL-4B, we construct a multimodal calibration set using 256 samples from the MathVista portion of the MathVerse dataset and evaluate on MMBench-en-dev (MMB) \cite{MMBench}, MMMU-val (MMMU) \cite{MMMU}, MMStar (MMS) \cite{MMStar}, OCRBench \cite{OCRBench}, and RealWorldQA (RWQA). In both cases, we compress the models by 20\% without any fine-tuning.

As shown in Table~\ref{tab:vlm_results}, ROCKET preserves strong performance on Qwen3-VL-4B, achieving 63.7 average accuracy (over 90\% of the original model's performance). On VibeVoice, Table~\ref{tab:speech} shows near-identical speech quality: WER remains stable (0.149 vs.\ 0.148), and UTMOS drops only slightly (3.43 vs.\ 3.52), staying close to the ground-truth reference (3.73). These results demonstrate that ROCKET generalizes effectively across modalities.

\begin{center}
  \begin{minipage}{0.52\columnwidth}
    \centering
    \captionof{table}{Average accuracy of Qwen3-VL-4B before and after ROCKET compression (20\%).}
    \label{tab:vlm_results}
    \resizebox{\linewidth}{!}{%
    \renewcommand{\arraystretch}{1.1}
    \begin{tabular}{lccccc}
    \toprule
    \textbf{Method} & \textbf{MMB} & \textbf{MMMU} & \textbf{MMS} & \textbf{OCR} & \textbf{RWQA} \\
    \midrule
    dense & 83.76 & 49.44 & 61.85 & 81.70 & 71.50 \\
    ROCKET     & 78.95 & 44.44 & 54.85 & 74.50 & 65.75 \\
    \bottomrule
    \end{tabular}
    }
  \end{minipage}%
  \hfill
  \begin{minipage}{0.39\columnwidth}
    \centering
    \captionof{table}{ROCKET applied to VibeVoice 7B at 20\% compression.}
    \label{tab:speech}
    \resizebox{\linewidth}{!}{%
    \renewcommand{\arraystretch}{1.1}
    \begin{tabular}{lcc}
    \toprule
    \textbf{Method} & \textbf{WER $\downarrow$} & \textbf{UTMOS $\uparrow$} \\
    \midrule
    Ground Truth    & 0.04  & 3.73 \\
    VibeVoice (dense) & 0.148 & 3.52 \\
    ROCKET          & 0.149 & 3.43 \\
    \bottomrule
    \end{tabular}
    }
  \end{minipage}
\end{center}

\vspace{-0.5 cm}
\section{Ablations}
\label{sec:ablations}

To evaluate the contribution of each key component in ROCKET, we conduct ablation studies on Llama3.2-1B following the evaluation protocol in Section~\ref{sec:exp_setup}. All variants are evaluated at a fixed global compression ratio of 20\%, with all other design choices held constant, and we report average accuracy along with word-level WikiText perplexity. Ablations on the reconstruction error metric (Appendix~\ref{app:recontruvtion_error_ablations}) and calibration data (Appendix~\ref{data_ablation}), per-benchmark results, and information about environmental impact and inference (Appendix~\ref{appendix:inference_optimization}) are provided in the appendix.

\begin{table}[t]
\centering
\setlength{\tabcolsep}{4pt}
\begin{minipage}[b]{0.43\linewidth}
    \centering
    \caption{Core components. ROCKET\textsuperscript{†} uses uniform budget allocation.}
    \label{tab:ablation}
    \begin{adjustbox}{max width=\linewidth}\footnotesize
    \renewcommand{\arraystretch}{1.0}
    \begin{tabular}{l c c c c}
    \toprule
    \textbf{Method} & \textbf{Sparse} & \textbf{\makecell{Budg.\\Alloc.}} & \textbf{\makecell{Avg.\\Acc. $\uparrow$}} & \textbf{PPL $\downarrow$} \\
    \midrule
    Dense & — & — & 57.6 & 12 \\
    \midrule
    SVD-LLM & \ding{55} & \ding{55} & 37.6 & 170 \\
    CoSpaDi & \checkmark & \ding{55} & 42.7 & 64 \\
    ROCKET\textsuperscript{†} & \checkmark & \ding{55} & 45.4 & 27 \\
    ROCKET & \checkmark & \checkmark & \textbf{52.4} & \textbf{18} \\
    \bottomrule
    \end{tabular}
    \end{adjustbox}
\end{minipage}%
\hfill
\begin{minipage}[b]{0.53\linewidth}
    \centering
    \caption{Sparsification of $\m C$: ROCKET variants (top) and pruning-style criteria after SVD (bottom).}
    \label{tab:sparsification_main}
    \begin{adjustbox}{max width=\linewidth}\footnotesize
    \renewcommand{\arraystretch}{1.0}
    \begin{tabular}{l l c c}
    \toprule
    \textbf{Strategy} & \textbf{Inspired by} & \textbf{\makecell{Avg.\\Acc. $\uparrow$}} & \textbf{PPL $\downarrow$} \\
    \midrule
    Per-Row & -- & 45.6 & 32 \\
    Global & -- & 50.1 & 19 \\
    Whitened-only ($\lambda=0$) & -- & 52.0 & 19 \\
    \midrule
    Uniform modular & MoDeGPT~\cite{lin2025modegpt} & 50.1 & 20 \\
    Outlier-weighted & OWL/Wanda~\cite{yin2023owl} & 34.4 & 187 \\
    Heavy-tailed self-reg. & AlphaPruning~\cite{lu2024alphapruning} & 39.9 & 57 \\
    \midrule
    \textbf{Column-wise + dual-space} & \textbf{ROCKET} & \textbf{52.4} & \textbf{18} \\
    \bottomrule
    \end{tabular}
    \end{adjustbox}
\end{minipage}

\vspace{0.05cm}
\begin{minipage}[b]{0.50\linewidth}
    \centering
    \caption{Step size of the knapsack CR grid.}
    \label{tab:step_size_ablation}
    \begin{adjustbox}{max width=\linewidth}\footnotesize
    \renewcommand{\arraystretch}{1.0}
    \begin{tabular}{lccc}
    \toprule
    \textbf{Step Size} & \textbf{\makecell{Total\\Rec. Error}} & \textbf{\makecell{Avg.\\Rec. Error}} & \textbf{\makecell{Profiling\\Time (s)}} \\
    \midrule
    No dynamic & 37.88 & 0.338 & 0 \\
    0.20 & 37.74 & 0.3370 & 208 \\
    0.15 & 32.53 & 0.2904 & 228 \\
    0.10 & 31.70 & 0.2830 & 308 \\
    \textbf{0.05} & \textbf{31.57} & \textbf{0.2819} & \textbf{425} \\
    0.025 & 31.56 & 0.2818 & 524 \\
    0.01 & 31.54 & 0.2816 & 1359 \\
    \bottomrule
    \end{tabular}
    \end{adjustbox}
\end{minipage}%
\hfill
\begin{minipage}[b]{0.45\linewidth}
    \centering
    \caption{Balance parameter $\lambda$ in~\eqref{eq:imp_final} (WikiText PPL).}
    \label{tab:lambda_ablation}
    \begin{adjustbox}{max width=\linewidth}\footnotesize
    \renewcommand{\arraystretch}{1.0}
    \begin{tabular}{cc|cc}
    \toprule
    $\boldsymbol{\lambda}$ & \textbf{PPL $\downarrow$} & $\boldsymbol{\lambda}$ & \textbf{PPL $\downarrow$} \\
    \midrule
    0.0 & 17.83 & 0.6 & 17.96 \\
    0.1 & 17.83 & 0.7 & 17.95 \\
    0.2 & 17.75 & 0.8 & 18.20 \\
    0.3 & \textbf{17.71} & 0.9 & 18.52 \\
    0.4 & 17.72 & 1.0 & 19.06 \\
    \textbf{0.5} & 17.75 & & \\
    \bottomrule
    \end{tabular}
    \end{adjustbox}
\end{minipage}
\vspace{-0.4cm}
\end{table}

\paragraph{Core components}
\label{app:core_components}
To assess the contribution of ROCKET's key design choices, we compare four variants: (1) SVD-LLM, which uses neither sparsification nor dynamic budget allocation; (2) CoSpaDi, which uses K-SVD-based sparse dictionary learning; (3) \ours{} with uniform compression across layers; and (4) full \ours{}, which further incorporates knapsack-based budget allocation. As shown in Table~\ref{tab:ablation}, using sparsification improves both average accuracy and perplexity. Moreover, replacing CoSpaDi's iterative sparsification with our closed-form, activation-aware approach is not only much faster, but also improves average accuracy from 42.7 to 45.4 and reduces perplexity from 64 to 27. Employing budget allocation yields a further significant gain, reaching 52.4 average accuracy and 18 perplexity, demonstrating that both our sparsification strategy and global parameter allocation are critical to \ours{}'s performance.

\vspace{-0.2cm}
\paragraph{Sparsification strategy}
\label{app:coeff_sparsification_comparison}
We compare \ours{} against alternative approaches for sparsifying the coefficient matrix $\m C$ (Table~\ref{tab:sparsification_main}): (1) \textbf{Per-Row}, where importance scores are computed identically but sparsity is enforced independently per row (breaking the column-wise structure); (2) \textbf{Global}, which ignores structural constraints and zeros out the globally least-important entries; and (3) \textbf{Whitened-only}, which disables the original-space fidelity term by setting $\lambda=0$ in~\eqref{eq:imp_final}. We further apply truncated SVD and then sparsify $\m C$ using strategies inspired by recent pruning methods~\cite{lin2025modegpt, yin2023owl, lu2024alphapruning}, with \ours{}'s budget allocation used for all variants. \ours{}'s column-aware, dual-space importance scoring outperforms others, demonstrating the efficiency of both structural awareness and dual-space scoring; transplanting pruning heuristics directly to the coefficient matrix is suboptimal, as the dictionary and coefficients must be treated jointly (detailed results in Appendix~\ref{app:sparsity_strategy_details}).

\paragraph{Balance parameter $\lambda$}
\label{sec:lambda_ablation}
Table~\ref{tab:lambda_ablation} shows strong robustness to $\lambda \in [0,1]$, with performance degrading only at extremes ($\lambda \geq 0.8$); we adopt $\lambda=0.5$ as a default for simplicity and stability.

\paragraph{Search discretization}
\label{subsec:ablation_search}
The discretization step size governs the resolution of the compression ratio grid in the multi-choice knapsack allocator. Table~\ref{tab:step_size_ablation} reports the impact of varying the step size on total reconstruction error, average per-layer error, and profiling runtime. The results justify our default choice of 0.05, as finer steps yield diminishing error improvements ($<0.03$ in total reconstruction error) while increasing profiling time by up to $3.2\times$. The error plateaus at $\approx 0.282$ for step sizes $\leq 0.10$, confirming numerical stability and indicating that the allocation obtained over the discretized candidate set is near-optimal with respect to finer grids (candidate-set construction in Appendix~\ref{app:search_space}).

\vspace{-0.8 cm}

\section{Conclusion and Limitations}
\ours~introduces a fast, training-free compression framework that combines calibration-guided structured factorization with principled knapsack-based budget allocation. It achieves state-of-the-art fidelity, retaining over 90\% of original accuracy at 30\% compression without fine-tuning.
While the exact DP solver scales efficiently for dense architectures, exhaustive search becomes intensive for extreme-scale MoE models. We address this via beam search and Pareto pruning, enabling practical compression for 100K+ layers with minor optimality trade-offs (please see Appendix~\ref{subsec:cross_arch_eval}).
Moreover, support selection relies on a diagonal surrogate that discards the off-diagonal couplings of $\m G$ (Section~\ref{sec:thresholding}), and its approximation error is not characterized theoretically; the whitening also depends on the calibration data, although the observed variation is small (Appendix~\ref{data_ablation}).
Finally, we note that all factorization-based methods (including \ours) reduce theoretical FLOPs but may incur sequential overhead on standard hardware. While \ours{} already reaches 1.07--1.37$\times$ the serving throughput of the dense model (Appendix~\ref{app:serving_throughput_memory}), the sparse kernel we use (\textsc{Macko}) does not yet support dynamic batching, and unlocking full inference acceleration requires specialized sparse-dense kernels and tiling strategies -- a promising direction for future system-level optimization.

\paragraph{Acknowledgments}
This research was supported as part of the ITMO University Research Projects in AI Initiative.

\bibliography{iclr2027_conference}
\bibliographystyle{acm}

\newpage
\appendix
\onecolumn
\section{Ethical Statement and Broader Impact} 
ROCKET is a training-free compression method designed to improve the efficiency and accessibility of large language models without requiring additional data or extensive computational resources for fine-tuning. By enabling high-fidelity model compression with minimal environmental and economic cost, it supports more sustainable deployment of AI systems, particularly in resource constrained settings.  
The method does not introduce new data collection, human annotation, or model behaviors beyond those already present in the original pretrained model; thus, it neither amplifies nor mitigates existing biases in the base model. Users should remain vigilant about the ethical implications of the underlying model’s outputs, as ROCKET preserves its functional characteristics including potential biases or safety limitations. We encourage responsible deployment, including thorough evaluation and alignment measures when compressed models are used in real-world applications.
\section{Proposed Algorithm}
\label{algorithm_full}
In Algo.~\ref{alg:ours}, we present the proposed method in the form of a formal algorithm to facilitate a clear and systematic understanding of the method's complete pipeline construction and implementation.
\begin{algorithm}[h]
\caption{\ours: Training-Free Heterogeneous transformer Compression}
\label{alg:ours}
\begin{algorithmic}[1]
\REQUIRE Pre-trained transformer with linear layers $\{\m{W}^{(\ell)}\}_{\ell=1}^L$, output layer data $\m{X}_{\ell} \in \mathbb{R}^{N \times d_1}$ (using calibration data $\m{X} \in \mathbb{R}^{N \times d_1}$), global parameter budget $C_{\text{total}}$
\ENSURE Compressed model with factorization $\{\m{B}^{(\ell)}, \m{C}^{(\ell)}\}_{\ell=1}^L$

\FOR{each layer $\ell = 1$ to $L$}
    \STATE Compute whitening transform: $\m{L}_{\ell} \gets \mathrm{chol}(\m{X}^\top_{\ell} \m{X}_{\ell} + \epsilon \m{I})$
    \STATE $\m{W}_{L}^{\ell} \gets \m{L}_{\ell} \m{W}^{(\ell)}$
    \FOR{each candidate rank $k \in \mathcal{K}$ and sparsity ratio $s \in \mathcal{S}$}
        \STATE Compute the top-$k$ eigenvectors $\m{B}$ via eigendecomposition: 
        \[
        \m{W}_L^{\ell} \m{W}_L^{{\ell}\top} \approx \m{B} \boldsymbol{\Lambda}_k \m{B}^\top
        \]
        \STATE Form coefficient matrix: $\m{C} \gets \m{B}^\top \m{W}_L^{\ell}$
        \STATE Compute importance scores: $\mathrm{imp}_{ij} \gets |c_{ij}| \cdot \|\m{L}_\ell^{-1} \m{b}_i\|_2^{\lambda}$ \COMMENT{$\lambda = 0.5$}
        \STATE Over-sparsify $\m{C}$ to ratio $s + \gamma$, then reactivate top entries globally to reach exact sparsity $s$
        \STATE Solve for optimal left factor: $\m{D} \gets \arg\min_{\m{D}} \|\m{W}_L^{\ell} - \m{D} \m{C}_{\text{sparse}}\|_F^2 + \mu \|\m{D}\|_F^2$
        \STATE Reconstruct weight: $\widetilde{\m{W}}^{\ell} \gets \m{L}_\ell^{-1} \m{D} \m{C}_{\text{sparse}}$
        \STATE Record option: cost $c_{\ell,i} \gets \texttt{nnz}(\m{D}) + \texttt{nnz}(\m{C}_{\text{sparse}})$, error $e_{\ell,i}$
        \STATE Store $(c_{\ell,i}, e_{\ell,i})$ in $\mathcal{O}_\ell$
    \ENDFOR
\ENDFOR

\STATE Compute reference error $\bar{e}_{\text{ref}}$ from uniform compression baseline and $\min \alpha$ that admits a solution
\STATE Initialize DP table: $\text{DP}_0[0] \gets 0$, others $\gets \infty$
\STATE $\beta_{\text{scale}} = \text{param\_precision} / C_{\text{total}}$
\FOR{$\ell = 1$ to $L$}
    \FOR{each state $p$ in $\text{DP}_{\ell-1}$}
        \FOR{each option $i \in \mathcal{O}_\ell$}
            \IF{$e_{\ell,i} \leq \alpha \cdot \bar{e}_{\text{ref}}$}
                \STATE $p' \gets p + \lfloor \beta_{\text{scale}} c_{\ell,i} \rfloor$
                \STATE $\text{DP}_\ell[p'] \gets \min\big(\text{DP}_\ell[p'],\ \text{DP}_{\ell-1}[p] + e_{\ell,i}\big)$
            \ENDIF
        \ENDFOR
    \ENDFOR
    \STATE Prune dominated states in $\text{DP}_\ell$: remove $(p_1, \text{err}_1)$ if $\exists (p_2, \text{err}_2)$ with $p_2 < p_1$ and $\text{err}_2 \leq \text{err}_1$
\ENDFOR

\STATE Find $p^* = \arg\min_{p \leq \lfloor \beta_{\text{scale}} C_{\text{total}} \rfloor} \text{DP}_L[p]$
\STATE Backtrack to recover the selected per-layer choices $\{i_\ell^*\}_{\ell=1}^L$
\FOR{$\ell = 1$ to $L$}
    \STATE Assign $\m{D}^{(\ell)}_{\text{final}}, \m{C}^{(\ell)}_{\text{final}}$ from option $i_\ell^*$
\ENDFOR
\end{algorithmic}
\end{algorithm}

\subsection{Theoretical Comparison with Iterative Dictionary Learning}
\label{subsec:theoretical_comparison}
It is important to highlight that CoSpaDi relies on unconstrained dictionary estimation via iterative K-SVD/OMP optimization, which is non-convex and lacks global convergence guarantees. In contrast, \ours{} employs a \textit{two-stage closed-form estimator}. First, we apply the Eckart--Young--Mirsky theorem via eigendecomposition to obtain a globally optimal column-orthogonal dictionary $\m{B}_k$. For this basis, sparse coding admits an \textit{exact analytical solution} via hard thresholding, avoiding the greedy approximation errors inherent to OMP. Second, we relax orthogonality constraints and compute the global least-squares update $\m{D}^* = \m{W}_L \m{C}^\top (\m{C}\m{C}^\top)^{-1}$, following the Method of Optimal Directions (MOD). While K-SVD updates atoms sequentially (coordinate descent), \ours{} computes the unique global minimizer for the dictionary given fixed codes. This sequential closed-form design combines spectral initialization stability with convex optimality, preventing error propagation and local minima typical of alternating minimization schemes.
\section{Norm-Preserving Properties of Low-Rank Approximation and Sparsification}
\label{reconstruction_error_dis}
Let $\mathbf{W} \in \mathbb{R}^{d_1 \times d_2}$ with singular value decomposition (SVD)
\[
\mathbf{W} = \mathbf{U} \boldsymbol{\Sigma} \mathbf{V}^\top,
\]
where $\mathbf{U} \in \mathbb{R}^{d_1 \times r}$, $\mathbf{V} \in \mathbb{R}^{d_2 \times r}$ have orthonormal columns, $\boldsymbol{\Sigma} = \operatorname{diag}(\sigma_1, \dots, \sigma_r)$, $\sigma_1 \geq \cdots \geq \sigma_r > 0$, and $r = \operatorname{rank}(\mathbf{W})$.

\subsection{Relative Error of Rank-$k$ Truncated SVD}
The optimal rank-$k$ approximation ($1 \leq k \leq r$) is
\[
\mathbf{W}_k = \mathbf{U}_k \boldsymbol{\Sigma}_k \mathbf{V}_k^\top,
\]
with error given by the Eckart--Young--Mirsky theorem:
\[
\| \mathbf{W} - \mathbf{W}_k \|_F^2 = \sum_{i=k+1}^r \sigma_i^2.
\]
Since $\| \mathbf{W} \|_F^2 = \sum_{i=1}^r \sigma_i^2$, it follows that
\[
\frac{\| \mathbf{W} - \mathbf{W}_k \|_F}{\| \mathbf{W} \|_F}
= \sqrt{ \frac{ \sum_{i=k+1}^r \sigma_i^2 }{ \sum_{i=1}^r \sigma_i^2 } }
\leq 1,
\]
with equality iff $k = 0$.

\subsection{Sparsification of the Coefficients}
Let $\mathcal{T}: \mathbb{R}^{k \times d_2} \to \mathbb{R}^{k \times d_2}$ denote an entrywise hard-thresholding operator. For any matrix $\mathbf{M}$, $\mathcal{T}(\mathbf{M})$ retains entries whose magnitudes exceed a prescribed threshold and sets the remainder to zero. Consequently, the retained component $\mathcal{T}(\mathbf{M})$ and the pruned residual $\mathbf{M} - \mathcal{T}(\mathbf{M})$ have disjoint supports, which implies orthogonality under the Frobenius inner product:
\[
\langle \mathcal{T}(\mathbf{M}), \mathbf{M} - \mathcal{T}(\mathbf{M}) \rangle_F = 0.
\]
By the Pythagorean theorem for the Frobenius norm,
\[
\| \mathbf{M} \|_F^2 = \| \mathcal{T}(\mathbf{M}) \|_F^2 + \| \mathbf{M} - \mathcal{T}(\mathbf{M}) \|_F^2,
\]
which yields the fundamental bound $\| \mathbf{M} - \mathcal{T}(\mathbf{M}) \|_F \leq \| \mathbf{M} \|_F$.

Let $\mathbf{W}_k = \mathbf{B}_k \boldsymbol{\Sigma}_k \mathbf{V}_k^\top$ be the rank-$k$ approximation of $\mathbf{W}$, where $\mathbf{B}_k \in \mathbb{R}^{d_1 \times k}$ has orthonormal columns ($\mathbf{B}_k^\top \mathbf{B}_k = \mathbf{I}_k$). Define the sparsified reconstruction as
\[
\widetilde{\mathbf{W}} = \mathbf{B}_k \, \mathcal{T}(\boldsymbol{\Sigma}_k \mathbf{V}_k^\top).
\]
Decompose $\mathbf{W}$ into its rank-$k$ component and orthogonal residual: $\mathbf{W} = \mathbf{W}_k + \mathbf{W}_{\perp}$, with $\mathbf{W}_{\perp} \perp \mathbf{W}_k$. Since $\widetilde{\mathbf{W}}$ lies in the column space of $\mathbf{B}_k$, it follows that $\mathbf{W}_{\perp} \perp (\mathbf{W}_k - \widetilde{\mathbf{W}})$. Therefore, the total reconstruction error decomposes as
\[
\| \mathbf{W} - \widetilde{\mathbf{W}} \|_F^2 = \| \mathbf{W}_{\perp} \|_F^2 + \| \mathbf{W}_k - \widetilde{\mathbf{W}} \|_F^2.
\]
Because $\mathbf{B}_k$ has orthonormal columns, left-multiplication preserves the Frobenius norm:
\[
\| \mathbf{W}_k - \widetilde{\mathbf{W}} \|_F = \| \boldsymbol{\Sigma}_k \mathbf{V}_k^\top - \mathcal{T}(\boldsymbol{\Sigma}_k \mathbf{V}_k^\top) \|_F.
\]
Applying the Pythagorean bound to $\mathbf{M} = \boldsymbol{\Sigma}_k \mathbf{V}_k^\top$ gives $\| \mathbf{W}_k - \widetilde{\mathbf{W}} \|_F \leq \| \boldsymbol{\Sigma}_k \mathbf{V}_k^\top \|_F = \| \mathbf{W}_k \|_F$. Substituting this into the error decomposition yields
\[
\| \mathbf{W} - \widetilde{\mathbf{W}} \|_F^2 \leq \| \mathbf{W}_{\perp} \|_F^2 + \| \mathbf{W}_k \|_F^2 = \| \mathbf{W} \|_F^2.
\]
Taking square roots establishes the rigorous relative error bound
\[
\frac{\| \mathbf{W} - \widetilde{\mathbf{W}} \|_F}{\| \mathbf{W} \|_F} \leq 1.
\]
Equality holds if and only if $\mathcal{T}$ zeroes out all coefficients ($\widetilde{\mathbf{W}} = \mathbf{0}$). For any nontrivial threshold, the relative error is strictly bounded by $1$.

\subsection{Equivalence of Eigenvalue-Based Basis Construction and SVD in ROCKET}
\label{SVD_and_evd}
In ROCKET, the weight matrix $\mathbf{W} \in \mathbb{R}^{d_1 \times d_2}$ is generally rectangular and non-symmetric, so eigendecomposition cannot be applied directly to $\mathbf{W}$. Instead, compression operates on the whitened weight $\mathbf{W}_L = \mathbf{L} \mathbf{W}$, where $\mathbf{L} = \mathrm{chol}(\mathbf{X}^\top \mathbf{X})$ whitens the input activations using a small calibration set $\mathbf{X} \in \mathbb{R}^{N \times d_1}$. The method then computes the top-$k$ eigenvectors of the symmetric positive semi-definite matrix $\mathbf{W}_L \mathbf{W}_L^\top$. This eigendecomposition yields
\[
\mathbf{W}_L \mathbf{W}_L^\top = \mathbf{B}_k \boldsymbol{\Lambda}_k \mathbf{B}_k^\top,
\]
where $\mathbf{B}_k \in \mathbb{R}^{d_1 \times k}$ has orthonormal columns ($\mathbf{B}_k^\top \mathbf{B}_k = \mathbf{I}_k$) and $\boldsymbol{\Lambda}_k = \operatorname{diag}(\eta_1, \dots, \eta_k)$ with $\eta_i \geq 0$.

Let the compact singular value decomposition of $\mathbf{W}_L$ be
\[
\mathbf{W}_L = \mathbf{U} \boldsymbol{\Sigma} \mathbf{V}^\top.
\]
Then
\[
\mathbf{W}_L \mathbf{W}_L^\top = \mathbf{U} \boldsymbol{\Sigma}^2 \mathbf{U}^\top,
\]
which is precisely the eigenvalue decomposition of $\mathbf{W}_L \mathbf{W}_L^\top$. Therefore, the eigenbasis coincides exactly with the left singular vectors: $\mathbf{B}_k = \mathbf{U}_k$ and $\eta_i = \sigma_i^2$. The coefficient matrix is obtained via orthogonal projection:
\[
\mathbf{C}_k = \mathbf{B}_k^\top \mathbf{W}_L = \mathbf{U}_k^\top \mathbf{W}_L = \boldsymbol{\Sigma}_k \mathbf{V}_k^\top,
\]
which matches the right factor in the SVD. Hence, the reconstruction
\[
\mathbf{W}_L = \mathbf{B}_k \mathbf{C}_k
\]
is identical to the rank-$k$ truncated SVD of $\mathbf{W}_L$.

Sparsification is applied to $\mathbf{C}_k$ using an operator $\mathcal{T}: \mathbb{R}^{k \times d_2} \to \mathbb{R}^{k \times d_2}$ satisfying
\[
\| \mathcal{T}(\mathbf{C}_k) \|_F \leq \| \mathbf{C}_k \|_F.
\]
The sparsified approximation in the whitened space is defined as
\[
\widetilde{\mathbf{W}}_L = \mathbf{B}_k \, \mathcal{T}(\mathbf{C}_k).
\]
Since $\mathbf{B}_k$ has orthonormal columns, the Frobenius norm is preserved under multiplication:
\[
\| \widetilde{\mathbf{W}}_L \|_F = \| \mathcal{T}(\mathbf{C}_k) \|_F \leq \| \mathbf{C}_k \|_F < \| \mathbf{W}_L \|_F \quad \text{for } k < r.
\]
\textit{(Note: Strict inequality holds as truncation removes non-zero singular components.)} Consequently, the relative reconstruction error in the whitened space satisfies
\[
\frac{\| \mathbf{W}_L - \widetilde{\mathbf{W}}_L \|_F}{\| \mathbf{W}_L \|_F} \leq 1,
\]
with equality only in the degenerate case $\widetilde{\mathbf{W}}_L = \mathbf{0}$. Because the eigenbasis $\mathbf{B}_k$ is mathematically identical to the left singular vectors of $\mathbf{W}_L$, all norm-preserving properties and error bounds established for truncated SVD carry over verbatim.

Finally, the compressed weight in the original space is recovered as
\[
\widetilde{\mathbf{W}} = \mathbf{L}^{-1} \widetilde{\mathbf{W}}_L.
\]
Although $\mathbf{L}^{-1}$ is not orthogonal, the theoretical guarantees in the whitened space where the core approximation occurs remain intact. Thus, despite using the eigendecomposition of $\mathbf{W}_L \mathbf{W}_L^\top$ for computational efficiency, ROCKET achieves the same norm constraints and relative error bound ($\leq 1$) as classical SVD-based low-rank approximation followed by norm-non-increasing sparsification.

% =====================================================================
% INSERT AFTER \subsection{Equivalence of Eigenvalue-Based Basis...}
% (i.e., after \label{SVD_and_evd} and its content)
% =====================================================================

\subsection{Exact Thresholding for Orthonormal Bases}
\label{app:pythagoras}

The following proposition justifies the use of hard thresholding as the
exact sparse-coding solver when the dictionary is column-orthogonal.

\begin{proposition}[Pythagorean decomposition]
\label{prop:pythagoras}
Let $\m{B}\in\R^{d_1\times k}$ have orthonormal columns and let
$\m{C}=\m{B}^{\top}\m{W}_L$.  For any
$\m{S}\in\R^{k\times d_2}$,
\begin{equation}
  \norm{\m{W}_L-\m{B}\m{S}}{F}^{2}
  =\norm{\m{W}_L-\m{B}\m{C}}{F}^{2}
  +\norm{\m{C}-\m{S}}{F}^{2}.
  \label{eq:pythagoras}
\end{equation}
Consequently, for fixed $\m{B}$, minimizing the whitened objective
over $s$-sparse $\m{S}$ decouples across columns and is solved by
retaining, in each column, the $s$ coefficients of largest magnitude
$|c_{ij}|$ --- the classical best-$s$-term approximation property of
orthonormal bases~\cite{mallat1999wavelet,elad2010sparse}.
\end{proposition}

\begin{proof}
Write
$\m{W}_L-\m{B}\m{S}
=(\m{I}-\m{B}\m{B}^{\top})\m{W}_L+\m{B}(\m{C}-\m{S})$.
The two terms are orthogonal: the first lies in
$\operatorname{span}(\m{B})^{\perp}$ and the second in
$\operatorname{span}(\m{B})$.  Moreover
$\norm{\m{B}\m{M}}{F}=\norm{\m{M}}{F}$ for any $\m{M}$
because $\m{B}^{\top}\m{B}=\m{I}$.
\end{proof}

For general (non-orthogonal) dictionaries no such exact rule exists,
and one must resort to greedy pursuits such as
OMP~\cite{tropp2007signal}.  This is precisely the cost that
\ours{} eliminates by restricting the sparse-coding step to an
orthogonal basis.

% =====================================================================
\subsection{Weighted Sparse Approximation and the Diagonal Surrogate}
\label{app:weighted-sap}

Proposition~\ref{prop:pythagoras} shows that plain thresholding of
$|c_{ij}|$ is the exact solver of the whitened objective.  However,
the ultimate goal is to reconstruct $\m{W}$, not $\m{W}_L$: the
un-whitening transform $\m{L}^{-1}$ is generally non-orthogonal, so
two coefficients with identical whitened magnitudes may contribute
very differently to the final reconstruction error.

\paragraph{The exact quadratic form.}
For a candidate coefficient matrix $\m{S}$, the perturbation
$\m{B}(\m{C}-\m{S})$ maps back to the original space as
$\m{L}^{-1}\m{B}(\m{C}-\m{S})$, whose energy is
\begin{equation}
  \norm{\m{L}^{-1}\m{B}(\m{C}-\m{S})}{F}^{2}
  =\operatorname{tr}\!\bigl(
    (\m{C}-\m{S})^{\top}\m{G}(\m{C}-\m{S})\bigr),
  \qquad
  \m{G}:=\m{B}^{\top}\m{A}^{-1}\m{B},
  \label{eq:G_full}
\end{equation}
with diagonal entries
$G_{ii}=\norm{\m{L}^{-1}\m{b}_i}{2}^{2}$.
The full form~\eqref{eq:G_full} couples coefficient rows through the
off-diagonal entries
$G_{ii'}=\langle\m{L}^{-1}\m{b}_i,\m{L}^{-1}\m{b}_{i'}\rangle$;
optimizing under~\eqref{eq:G_full} directly would require
combinatorial search.

\paragraph{The diagonal surrogate.}
\ours{} replaces $\m{G}$ by the powered diagonal surrogate
\begin{equation}
  \m{G}\;\approx\;\operatorname{diag}(\m{G})^{\lambda},
  \qquad \lambda\in[0,1],
  \label{eq:surrogate}
\end{equation}
which renders the objective separable across entries.  The surrogate
has a precise meaning at the extremes:
\begin{itemize}
  \item $\lambda=0$: the surrogate is the identity, recovering the
    \emph{exact} whitened error of
    Proposition~\ref{prop:pythagoras};
  \item $\lambda=1$: the surrogate is the diagonal part of the
    \emph{exact} original-space quadratic
    form~\eqref{eq:G_full};
  \item $\lambda=0.5$: the geometric mean of the two error models.
\end{itemize}
Both neglected terms the off-diagonal couplings $G_{ii'}$ and
the truncation-residual coupling  vanish in the isotropic limit
$\m{A}\propto\m{I}$, where $\m{G}=\m{I}$ and the whitened and
original geometries coincide.  Empirically, the surrogate is
insensitive to $\lambda$ over a wide range
(Section~\ref{sec:lambda_ablation}), and the dual-space scoring
consistently outperforms whitened-only scoring and transplanted
pruning heuristics
(Section~\ref{app:coeff_sparsification_comparison}).

\paragraph{Weighted best-subset thresholding.}
With $\nu_i:=G_{ii}^{\lambda/2}
=\norm{\m{L}^{-1}\m{b}_i}{2}^{\lambda}$, support selection under
the surrogate amounts to the weighted sparse approximation problem
\begin{equation}
  \m{S}^{\star}\in\argmin_{\m{S}}\;
  \norm{\operatorname{diag}(\boldsymbol{\nu})
  (\m{C}-\m{S})}{F}^{2}
  \quad\text{s.t.}\quad
  \operatorname{supp}(\m{S})\in\mathcal{F},
  \label{eq:wsap}
\end{equation}
where $\mathcal{F}$ encodes the per-column and global sparsity
constraints.  For a fixed support $\Omega$, the optimal coefficients
are $S^{\star}_{ij}=C_{ij}$ on $\Omega$ and zero elsewhere, and the
objective reduces to
\begin{equation}
  \text{const.}
  \;-\;\sum_{(i,j)\in\Omega}\nu_i^{2}\,c_{ij}^{2}
  =\text{const.}
  \;-\;\sum_{(i,j)\in\Omega}\mathrm{imp}_{ij}^{2},
  \qquad
  \mathrm{imp}_{ij}:=\nu_i\,|c_{ij}|.
  \label{eq:imp_equiv}
\end{equation}
Support selection therefore reduces to maximizing the accumulated
importance $\sum_{(i,j)\in\Omega}\mathrm{imp}_{ij}^{2}$.

\begin{lemma}[Weighted best-subset thresholding]
\label{lem:thresholding}
Let $\nu_i>0$ and $w_{ij}:=\nu_i^{2}c_{ij}^{2}\ge 0$.
\begin{enumerate}
  \item Under a per-column cardinality constraint
    $|\Omega_j|=s$, the optimal support keeps the $s$ largest
    scores $\nu_i|c_{ij}|$ in each column.
  \item Under the mixed constraint $|\Omega_j|\ge s_0$ for all $j$
    and $|\Omega|=K$ (with $K\ge d_2 s_0$), an optimal support is
    obtained by first keeping the top-$s_0$ entries in each column
    and then reactivating the globally largest remaining scores
    until $K$ entries are selected.
\end{enumerate}
\end{lemma}

\begin{proof}
(i)~The objective and constraint decouple across columns; within a
column, keeping the $s$ largest weights is optimal.

(ii)~Let $T_j$ denote the top-$s_0$ entries of column $j$ and
$T=\bigcup_j T_j$.  An exchange argument shows that some optimal
solution contains $T$: if $i\in T_j\setminus\Omega$ with
$|\Omega_j|>s_0$, swapping $i$ with a lowest-weight element of
$\Omega_j\setminus T_j$ preserves feasibility and does not decrease
the objective.  Conditioned on $\Omega\supseteq T$, all column-wise
lower bounds are satisfied, so the remaining $K-d_2 s_0$ selections
are unconstrained and must be the globally largest remaining weights.
\end{proof}

Rule~(ii) of Lemma~\ref{lem:thresholding} is exactly \ours{}'s
two-stage thresholding procedure: over-sparsify per column to
$\mathrm{cr}+\gamma$, then reactivate globally top entries until the
target CR is reached.

% =====================================================================
\subsection{Recovery of Standard SVD}
\label{app:svd-recovery}

We verify that \ours{} reduces to standard data-aware low-rank SVD
when no sparsity is applied.

\paragraph{Case $s=k$ (no sparsity).}
When the column sparsity equals the dictionary size ($s=k$,
equivalently $k/s=1$), no coefficient is removed and
$\m{C}_{\mathrm{sparse}}=\m{C}=\boldsymbol{\Sigma}_k\m{V}_k^{\top}$.
Taking $\mu\to 0$ in the ridge-regularized least-squares
problem~\eqref{eq:ls}, the MOD update gives
\begin{equation}
  \m{D}_{\mathrm{final}}
  =\m{W}_L\m{C}^{\top}(\m{C}\m{C}^{\top})^{-1}
  =\m{U}_k\boldsymbol{\Sigma}_k^{2}
   (\boldsymbol{\Sigma}_k^{2})^{-1}
  =\m{U}_k,
\end{equation}
so that
$\widetilde{\m{W}}_L
=\m{D}_{\mathrm{final}}\m{C}
=\m{U}_k\boldsymbol{\Sigma}_k\m{V}_k^{\top}$,
which is exactly the rank-$k$ truncated SVD of $\m{W}_L$.
Empirically, on Llama3.2-1B at CR $=0.2$, \ours{} with $k/s=1$ and uniform allocation reproduces SVD-LLM exactly (37.6 average accuracy and 170 WikiText perplexity).

With the ridge term included,
$\m{D}_{\mathrm{final}}
=\m{U}_k\boldsymbol{\Sigma}_k^{2}
(\boldsymbol{\Sigma}_k^{2}+\mu\m{I})^{-1}$,
which approaches $\m{U}_k$ as $\mu\to 0$.  In \ours{}, $\mu$ is
adaptive and negligible relative to the matrix scale
(Appendix~\ref{hyp_ablation_details}).

\paragraph{Case $s<k$ (union of subspaces).}
Each column retains a different subset of $s$ coefficients, so each
reconstructed column lies in a different $s$-dimensional subspace.
The model therefore represents a union of subspaces, strictly more
expressive than the single-subspace rank-$k$ SVD.
\section{Mapping MCKP to Graph Theory.}
\label{graphtheorymckp}
We reformulate the constrained multi-choice knapsack problem (MCKP) as a shortest-path problem on a directed acyclic graph (DAG), where the key to enforcing the global parameters kept  lies in a clever \textit{target-aware sink connectivity rule}. Each node is labeled $(\ell, p)$, with $\ell \in \{0, \dots, L\}$ denoting the layer index and $p \in \mathbb{Z}_{\geq 0}$ representing the scaled cumulative number of parameters kept up to layer $\ell$, discretized via $p = \big\lfloor \texttt{scale} \cdot \sum_{\ell'=1}^{\ell} c_{\ell',i} \big\rfloor$ using $\texttt{scale} = \texttt{param\_precision} / C_{\text{total}}$. For each feasible compression option $i$ in layer $\ell$, we add an edge from $(\ell{-}1, p_{\text{in}})$ to $(\ell, p_{\text{out}})$ with $p_{\text{out}} = p_{\text{in}} + \lfloor \texttt{scale} \cdot c_{\ell,i} \rfloor$ and edge cost $\lfloor \texttt{error\_scale\_factor} \cdot e_{\ell,i} \rfloor$, while discarding any option violating the per-layer error cap $e_{\ell,i} \leq \alpha \bar{e}_{\text{ref}}$. Crucially, we connect a terminal node $(L, p)$ to the sink \textit{if and only if} $p \leq p_{\max} = \lfloor \texttt{scale} \cdot C_{\text{total}} \rfloor$, where $C_{\text{total}} = (1 - \rho_{\text{target}})P_{\text{orig}}$ is the kept-parameter budget and $\rho_{\text{target}}$ is the desired compression ratio (fraction of parameters removed). This structural embedding of the global budget constraint via node naming and selective sink connectivity ensures that any path reaching the sink automatically satisfies the target compression, transforming the constrained combinatorial optimization into an unconstrained shortest-path search solvable exactly (up to discretization) by Dijkstra’s algorithm. Since the MCKP is NP-hard, this exact solution is pseudo-polynomial in the number of budget bins; dominated-state (Pareto) pruning retains only non-dominated (cost, error) states, which keeps the search tractable in practice (Appendix~\ref{sec:scalability_runtime}). The graph thus encodes both layer-wise flexibility and global feasibility, with complexity rendered tractable by limiting the candidate space per layer to a modest grid of compression ratios and $k_s$ sparsity levels.

\begin{figure}[h]
\centering
\resizebox{0.5\textwidth}{!}{%  % Resize to full text width; keep aspect ratio
\begin{tikzpicture}[
    node/.style={draw, circle, minimum size=20pt, inner sep=0pt, font=\small},
    source/.style={fill=blue!20},
    dpnode/.style={fill=green!15},
    sink/.style={fill=red!20},
    edge/.style={->, >=stealth, thin, gray!60},
    goodedge/.style={->, >=stealth, red, thick},
    every node/.style={transform shape}
]

% Source: processed 0 layers, kept = 0
\node[node, source] (s) at (0,0) {$(0, 0)$};
\node[left=8pt of s, font=\small] {Start (0 layers)};

% After choosing option for layer 0 (step 1)
\node[node, dpnode] (n1a) at (2.5,-1.5) {$(1, p^{(1)}_1)$};
\node[node, dpnode] (n1b) at (2.5,0)    {$(1, p^{(2)}_1)$};
\node[node, dpnode] (n1c) at (2.5,1.5)  {$(1, p^{(3)}_1)$};
\node[below=40pt of n1b, font=\footnotesize] {After layer 0};

% After choosing option for layer 1 (step 2)
\node[node, dpnode] (n2a) at (5,-2)   {$(2, p^{(1)}_2)$};
\node[node, dpnode] (n2b) at (5,-0.5) {$(2, p^{(2)}_2)$};
\node[node, dpnode] (n2c) at (5,1)    {$(2, p^{(3)}_2)$};
\node[node, dpnode] (n2d) at (5,2.5)  {$(2, p^{(4)}_2)$};
\node[below=40pt of n2b, font=\footnotesize] {After layer 1};

% After choosing option for layer L-1 (step L)
\node[node, dpnode] (nLa) at (7.5,-1.5) {$(L, p^{(1)}_L)$};
\node[node, dpnode] (nLb) at (7.5,0)    {$(L, p^{(2)}_L)$};
\node[node, dpnode] (nLc) at (7.5,1.5)  {$(L, p^{(3)}_L)$};
\node[below=40pt of nLb, font=\footnotesize] {After layer $L{-}1$};

% Sink: only if total kept >= p_min
\node[node, sink] (t) at (10,0) {$(L, p \leq p_{\max})$};
\node[above=2pt of t, font=\footnotesize] {Feasible sink};

% Edges: step 0 → step 1 (using layer 0 options)
\draw[edge] (s) -- (n1a);
\draw[edge] (s) -- (n1b);
\draw[edge] (s) -- (n1c);

% Edges: step 1 → step 2 (using layer 1 options)
\foreach \from in {n1a,n1b,n1c} {
    \foreach \to in {n2a,n2b,n2c,n2d} {
        \draw[edge] (\from) -- (\to);
    }
}

% Edges: step 2 → step L (using layer L-1 options)
\foreach \from in {n2a,n2b,n2c,n2d} {
    \foreach \to in {nLa,nLb,nLc} {
        \draw[edge] (\from) -- (\to);
    }
}

% Only connect to sink if p >= p_min
\draw[goodedge] (nLb) -- (t);
\draw[goodedge] (nLc) -- (t);

% p_min threshold
\draw[dashed, gray] (9.2,-3) -- (9.2,2.5)
    node[above, black, font=\scriptsize] {$p = p_{\max}$};

% Callout
\node[align=left, font=\scriptsize, text width=5.5cm, draw=gray!50, rounded corners=2pt] (eq) at (12.5,-2.2) {
    State $(i, p)$: \\
    \quad $i =$ number of layers processed \\
    \quad $p =$ scaled total parameters kept \\
    Transition: pick option for layer $i$ \\
    Sink: only if $p \leq p_{\max}$
};
\draw[->, thin, gray] (eq.west) -- ++(-0.5,0) -- ++(0,-1.4) -- (nLc.east);

\end{tikzpicture}
}
\caption{Exact state-space graph matching. Each state $(i, p)$ represents having processed the first $i$ layers with $p$ scaled parameters retained. From $(i, p)$, the algorithm branches to all options for layer $i$, producing states $(i+1, p + \Delta p)$. The sink is reachable only from states with $p \leq p_{\max}$, enforcing the global compression ratio by construction.}
\label{fig:graph_\ours_dp}
\end{figure}
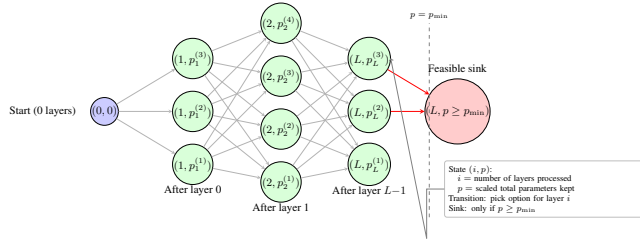
This graph formulation underlies the dynamic programming procedure in Algorithm \ref{alg:ours}, where states correspond to nodes and transitions to edges.
\section{Inference Optimization}
\label{appendix:inference_optimization}
To accelerate inference, we leverage \textsc{Macko}~\cite{macko}, a specialized sparse matrix--vector multiplication kernel that outperforms PyTorch’s built-in implementation for structured, column-wise sparse matrices. Because ROCKET employs a calibration-guided, adaptive compression strategy, the sparsity level quantified by the ratio \(k/s\) of dictionary atoms to nonzeros per coefficient column varies across layers. This heterogeneity arises from the solution to a constrained multi-choice knapsack problem, which allocates parameters to layers based on their marginal contribution to reconstruction fidelity under a global budget.

Empirically, we observe that MLP layers (gate and up projections) are assigned significantly higher sparsity and lower rank than attention projections (query, key, value, out). This behavior is theoretically justified: MLP weight matrices are substantially larger (\(d_{\text{model}} \times 4d_{\text{model}}\)) than attention matrices (\(d_{\text{model}} \times d_{\text{model}}\)), and their calibration-aware reconstruction error increases more slowly with sparsification. Consequently, the optimizer preferentially compresses MLP layers to maximize parameter savings while preserving overall model accuracy.

\begin{figure}[h]
    \centering
    \includegraphics[width=1.0\linewidth]{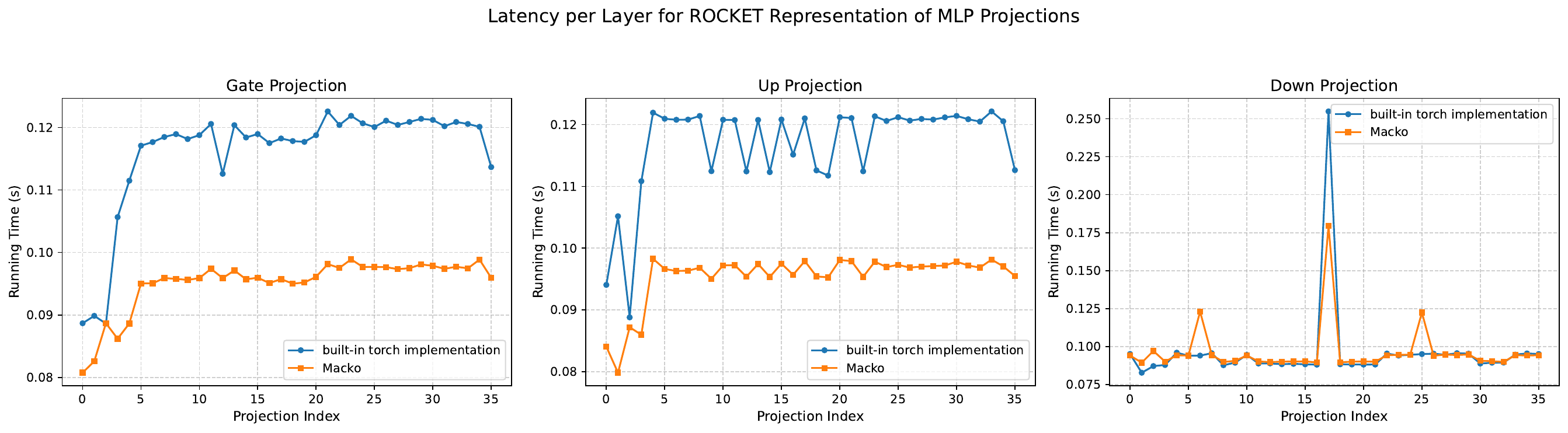}
    \caption{Comparison between PyTorch’s built-in sparse-matrix vector multiplication and \textsc{Macko} across MLP layers in Qwen3-8B. \textsc{Macko} shows consistently better running time for large coefficients (gate and up projections), while being on par with PyTorch for smaller sparse matrices (down projection).}
    \label{runningtime}
\end{figure}

Figure~\ref{runningtime} compares the runtime of MLP projections in Qwen3-8B using either PyTorch’s default sparse kernels or \textsc{Macko}. For large, moderately sparse coefficient matrices (e.g., gate and up projections), \textsc{Macko} provides consistent speedups; for smaller or denser matrices (e.g., down projections), both implementations perform similarly. In attention layers, where weight matrices are small and less aggressively compressed, we retain PyTorch’s native implementation, as it proves faster in practice. Additionally, we fuse dictionaries and coefficients for layers sharing the same input namely, \{query, key, value\} in attention and \{gate, up\} in the MLP  before applying \textsc{Macko}.

In terms of theoretical floating-point operations (FLOPs), ROCKET is structurally analogous to CoSpaDi: both represent a weight matrix \(\mathbf{W} \in \mathbb{R}^{d_1 \times d_2}\) as \(\mathbf{W} \approx \mathbf{B} \mathbf{C}\), where \(\mathbf{B} \in \mathbb{R}^{d_1 \times k}\) is a dense dictionary and \(\mathbf{C} \in \mathbb{R}^{k \times d_2}\) is column-wise sparse. Assuming optimal reuse of the intermediate product \(\mathbf{X}\mathbf{D}\), the FLOP count is \(N d_1 K_{\text{active}} + N s d_2\), where \(K_{\text{active}}\) is the number of distinct active atoms. Under a fixed global parameter budget, the total FLOP count may be similar between the two methods.

However, a key distinction lies in budget allocation: CoSpaDi typically enforces uniform sparsity across layers, whereas ROCKET dynamically assigns heterogeneous \((k_\ell, s_\ell)\) pairs per layer based on reconstruction sensitivity. This results in lower FLOPs in large, robust layers (e.g., MLP) and higher FLOPs in small, sensitive layers (e.g., attention) even though the latter contribute minimally to total computation due to their size. The net effect is a more balanced per-layer runtime profile and better utilization of sparse kernels, which explains the consistent throughput advantage of ROCKET over CoSpaDi (Table~\ref{tab:throughput}) despite comparable theoretical operation counts.

\begin{table}[h]
\centering
\caption{Throughput (tokens/s) for Qwen3-8B with batch\_size = 1 and context\_length = 256.}
\label{tab:throughput}
\begin{tabular}{lccc}
\toprule
Method & 20\% & 30\% & 40\% \\
\midrule
SVD-LLM & 24.36 & 24.31 & 24.72 \\
CoSpaDi & 25.45 & 25.76 & 25.62 \\
ROCKET  & 26.74 & 26.60 & 26.36 \\
\bottomrule
\end{tabular}
\end{table}

\subsection{Environmental Impact}

As shown in Table~\ref{tab:energy_emission}, the \textsc{\ours} method not only achieves superior compression performance but also offers dramatic environmental benefits. Compared to \textsc{CoSpaDi}, \textsc{\ours} consumes over 100 times less energy, completes compression 96 times faster, and produces 23 times lower CO\textsubscript{2} emissions. These results highlight \textsc{\ours} as both a high-performance and environmentally sustainable solution.
\begin{table}[h]
\centering
\caption{Comparison of energy consumption, runtime, and CO\textsubscript{2} emissions for CoSpaDi and ROCKET using the Llama3.2-1B model.}
\label{tab:energy_emission}
\resizebox{1.0\textwidth}{!}{
\begin{tabular}{lcccccc}
\toprule
Method & Model & GPU & CPU & Energy Consumed (kWh) & Duration (s) & CO\textsubscript{2} Emissions (kg eq) \\
\midrule
CoSpaDi & Llama3.2-1B & 1 × NVIDIA A100-SXM4-40GB & AMD EPYC 7742 64-Core Processor & 7.88 & 90080.97 & 0.782 \\
ROCKET  & Llama3.2-1B & 1 × NVIDIA A100-SXM4-40GB & AMD EPYC 7742 64-Core Processor & 0.0765 & 930 & 0.0337 \\
\bottomrule
\end{tabular}
}
\end{table}
\subsection{Hardware Efficiency and Sparse Kernel Compatibility}
\label{subsec:hardware_efficiency}
Recent advances in sparse linear algebra have made heterogeneous sparsity increasingly hardware-friendly. We leverage \textsc{Macko}~\cite{macko}, a specialized kernel optimized for column-wise sparse matrices, which achieves \textbf{1.3--1.5$\times$} throughput gains over cuBLAS at 50\% density and \textbf{3.5--4.5$\times$} at 10\% density. As shown in Table~\ref{tab:throughput}, \ours{} consistently outperforms uniform-compression baselines under matched budgets.

While \textsc{Macko} currently lacks dynamic batching support (planned for future releases), ROCKET's structural advantages persist with standard PyTorch sparse kernels. The FLOP count for a forward pass is $N(d_1 k_{\text{active}} + s d_2)$, where $k_{\text{active}}$ denotes the number of distinct active dictionary atoms. Our knapsack allocator preferentially assigns higher sparsity to large, robust layers (e.g., MLP projections), effectively reducing $k_{\text{active}}$ and $s$ in compute-heavy components. Empirically, this yields \textbf{5--12\% lower end-to-end latency} compared to low-rank baselines even without \textsc{Macko}, confirming that heterogeneous allocation intrinsically improves hardware utilization.

\subsection{Serving-Style Throughput and Memory Footprint}
\label{app:serving_throughput_memory}

To assess ROCKET under realistic serving workloads, we measure total system throughput under concurrent users, following CoSpaDi's evaluation protocol. All throughput values are reported as relative tokens/s with respect to the dense model; values greater than 1.00 indicate higher throughput than the uncompressed model.

\paragraph{Short-prompt throughput.}
Table~\ref{tab:short_prompt_throughput} reports relative throughput at prompt length $= 1$ under varying concurrency. ROCKET achieves the highest throughput at 1, 4, and 16 concurrent users and remains competitive at 8 users.

\begin{table}[h]
\centering
\caption{Short-prompt serving throughput (prompt length $= 1$), relative tokens/s vs.\ dense model.}
\label{tab:short_prompt_throughput}
\begin{tabular}{lcccc}
\hline
Method & 1 user & 4 users & 8 users & 16 users \\
\hline
SVD-LLM & 1.29 & 1.33 & \textbf{1.35} & 0.99 \\
CoSpaDi & 1.24 & 1.30 & 1.29 & 1.13 \\
ROCKET  & \textbf{1.31} & \textbf{1.37} & 1.31 & \textbf{1.14} \\
\hline
\end{tabular}
\end{table}

\paragraph{Prompt-length sweep.}
Table~\ref{tab:prompt_sweep_throughput} reports relative throughput across prompt lengths and concurrency levels (format: CoSpaDi / ROCKET). ROCKET matches or exceeds CoSpaDi in most configurations.

\begin{table}[h]
\centering
\caption{Full prompt-length sweep, relative tokens/s vs.\ dense (format: CoSpaDi / ROCKET).}
\label{tab:prompt_sweep_throughput}

\begin{tabular}{lccccc}
\hline
Users & Prompt 1 & Prompt 16 & Prompt 32 & Prompt 64 & Prompt 128 \\
\hline
1  & 1.24 / \textbf{1.31} & 1.11 / \textbf{1.19} & 1.09 / \textbf{1.15} & \textbf{1.20} / 1.16 & \textbf{1.14} / 1.12 \\
4  & 1.30 / \textbf{1.37} & 1.09 / \textbf{1.19} & \textbf{1.11} / 1.08 & \textbf{1.22} / 1.20 & \textbf{1.18} / 1.12 \\
8  & 1.29 / \textbf{1.31} & \textbf{1.08} / 1.07 & 1.11 / \textbf{1.14} & 1.24 / \textbf{1.25} & 1.15 / 1.15 \\
16 & 1.13 / \textbf{1.14} & 1.09 / \textbf{1.14} & 1.08 / \textbf{1.15} & 1.13 / \textbf{1.16} & 1.19 / 1.16 \\
\hline
\end{tabular}

\end{table}

These results show that ROCKET is not only accurate but also practically competitive under realistic serving workloads. We note that MACKO currently lacks dynamic batching support; throughput gains are expected to increase once sparse-kernel batching is available.

\paragraph{Memory footprint.}
Table~\ref{tab:memory_footprint} summarizes the stored parameter count for a single layer $W \in \mathbb{R}^{d_1 \times d_2}$ under different compression strategies.

\begin{table}[h]
\centering
\caption{Per-layer memory footprint comparison for $W \in \mathbb{R}^{d_1 \times d_2}$.}
\label{tab:memory_footprint}
\begin{tabular}{lc}
\hline
Method & Stored Parameters \\
\hline
Dense & $d_1 \cdot d_2$ \\
Rank-$k$ SVD & $k(d_1 + d_2)$ \\
ROCKET & $d_1 \cdot k + s \cdot d_2$ \\
\hline
\end{tabular}
\end{table}

ROCKET stores a dense factor $\m F \in \mathbb{R}^{d_1 \times k}$ and a column-wise sparse factor $\m H = \m C_{\text{sparse}} \in \mathbb{R}^{k \times d_2}$ with $s$ nonzeros per column. Sparse indices for $\m H$ are accounted for by truncating the float16 representation by 2 bits, following CoSpaDi's convention.

\section{Scalability and Runtime Profiling}
\label{sec:scalability_runtime}
The computational overhead of \ours{} consists of three components: activation gathering, whitening, and the DP allocation search. 

\paragraph{DP Allocation Cost.} 
For each linear module, we compute a single SVD/eigendecomposition. Subsequent rank/sparsity evaluations rely on pointer traversal across singular values and closed-form matrix operations, making the DP search complexity independent of weight matrix dimensions. As shown in Table~\ref{tab:dp_scaling}, DP search runtime scales super-linearly in practice (empirically $\approx O(L^{1.8})$) due to Pareto pruning overhead and state management.
\begin{center}
  \begin{minipage}{0.4\columnwidth}
    \centering
    \captionof{table}{DP search runtime on a single CPU thread.}
    \label{tab:dp_scaling}
    \resizebox{\linewidth}{!}{%
    \renewcommand{\arraystretch}{1.1}
    \begin{tabular}{ccc}
    \toprule
    \textbf{Layers} & \textbf{Options} & \textbf{Run Time (s)} \\
    \midrule
    112 & 14 & 9.80 \\
    224 & 14 & 35.00 \\
    336 & 14 & 69.75 \\
    448 & 14 & 115.93 \\
    \bottomrule
    \end{tabular}
    }
  \end{minipage}
    \hfill
  \begin{minipage}{0.54\columnwidth}
    \centering
    \captionof{table}{Whitening runtime for varying hidden dimensions (1024 calibration tokens).}
    \label{tab:whitening_time}
    \resizebox{\linewidth}{!}{%
    \renewcommand{\arraystretch}{1.1}
    \begin{tabular}{ccc}
    \toprule
    \textbf{Hidden Dim} & \textbf{$X^\top X$ + Cholesky} & \textbf{Time (s)} \\
    \midrule
    100 & $100 \times 100$ & 0.0005 \\
    1,000 & $1000 \times 1000$ & 0.0009 \\
    10,000 & $10000 \times 10000$ & 0.0450 \\
    28,672 & $28672 \times 28672$ & 0.3800 \\
    50,000 & $50000 \times 50000$ & 3.9600 \\
    \bottomrule
    \end{tabular}
    }
  \end{minipage}
\end{center}
Even for dense 70B-parameter models (448 linear layers), the solver completes in $<2$ minutes on a single CPU thread.\\
\textbf{Whitening Overhead.} 
Cholesky-based whitening ($\m{L} = \mathrm{chol}(\m{X}^\top \m{X})$) is performed once per layer using a small calibration set. Table~\ref{tab:whitening_time} demonstrates that this step is negligible compared to SVD/factorization costs.
\section{Further results}
\label{more_results}
\subsection{Detailed Comparison with CoSpaDi and SVD-LLM}
\label{detailed_comparison}
We begin this section by providing the detailed per-benchmark results from Figure~\ref{othmodels} in Tables \ref{tab:mainresults_detailed}, \ref{tab:main_llama3_detailed}. As shown, \ours~ outperforms CoSpaDi and SVD-LLM by a large margin across different benchmarks and compression ratios.
\begin{table}[H]
\caption{\ours{} comparison vs low-rank and SDL counterparts in data-aware scenarios on Llama3.2-1B at different compression ratios (CR). Best results are provided in \textbf{bold} All experiments are in training-free setup.}
\label{tab:mainresults_detailed}
\resizebox{\textwidth}{!}{%
\renewcommand{\arraystretch}{1.35}
\begin{tabular}{ccccccccccc|cc}
\hline
                           &                       & \multicolumn{9}{c}{\textbf{Accuracy$\uparrow$}}    & \multicolumn{2}{c}{\textbf{Perplexity$\downarrow$}} \\ \cline{3-12}\cline{12-13}
                          
\multirow{-2}{*}{\textbf{Method}} & \multirow{-2}{*}{\textbf{CR}}  & \textbf{PIQA} & \textbf{Hella Swag} & \textbf{LAMBADA} & \textbf{ARC-e} & \textbf{ARC-c} & \textbf{SciQ} & \textbf{Race}& \textbf{MMLU} & \textbf{Avg.} & \textbf{Wiki Text} & \textbf{LAMBADA}    \\ \hline
\textbf{Llama3.2 1B}  & --              & 74.5 & 63.7 & 63.0 & 60.5 & 36.2 & 88.3 & 37.8 & 37.0 & 57.6 & 12 & 5.7 \\ \hline
SVD-LLM     &                       & 62.1 & 36.4 & 24.4 & 36.0 & 25.1 & 64.9 & 29.0 & 23.0 & 37.6          & 170 & 170          \\
CoSpaDi     &  & 66.1 & 42.9 & 38.4 & 39.9 & 26.0 & 71.6 & 31.7 & 24.8 & 42.7 & 64 & 35 \\
\ours    & \multirow{-3}{*}{0.2} & \textbf{71.9} & \textbf{56.7} & \textbf{46} & \textbf{56.7} & \textbf{32.4} & \textbf{88.6} & \textbf{36.4} & \textbf{30.8} & \textbf{52.4} & \textbf{18} & \textbf{13} \\ \hline

SVD-LLM     &                       & 55.7 & 30.1 & 9.1  & 30.5 & 21.5 & 45.9 & 25.8 & 23.2 & 30.2          & 590          & 2,500          \\
CoSpaDi     &  & 56.9 & 32.4 & 18.2 & 31.9 & 22.1 & 56.7 & 28.0 & 23.1 & 33.7 & 290 & 660 \\
\ours     & \multirow{-3}{*}{0.3} & \textbf{66.3} & \textbf{46.8} & \textbf{38.0} & \textbf{47.9} & \textbf{27.4} & \textbf{79.6} & \textbf{34.1} & \textbf{27.4} & \textbf{45.9} & \textbf{35} & \textbf{26} \\ \hline

SVD-LLM     &                       & 51.8 & 27.3 & 1.3  & 26.9 & 22.9 & 32.3 & 24.4 & 23.0 & 26.2          & 1,600          & 33,000          \\
CoSpaDi     &  & 53.5 & 28.2 & 3.8  & 27.8 & 23.0 & 36.9 & 24.0 & \textbf{23.1} & 27.5 & 800 & 9,200 \\
\ours    & \multirow{-3}{*}{0.4} & \textbf{63.9} & \textbf{39.4} & \textbf{23.8}  & \textbf{41.0} & \textbf{25.7} & \textbf{72.1} & \textbf{30.9} & 23.0 & \textbf{39.8} & \textbf{88} & \textbf{130} \\ \hline

SVD-LLM     &                       & 51.1 & 26.6 & 0.0  & 26.1 & \textbf{25.9} & 26.1 & 23.9 & 23.0 & 25.4          & 3,100          & 100,000          \\
CoSpaDi     &  & 51.7 & 27.0 & 0.3  & 26.3 & 24.0 & 29.5 & 24.2 & \textbf{23.3} & 25.8 & 1,800 & 73,000 \\
\ours     & \multirow{-3}{*}{0.5} & \textbf{57.3} & \textbf{31.6} & \textbf{9.4}  & \textbf{34.9} & 22.6 & \textbf{50} & \textbf{26.1} & 22.9 & \textbf{31.9} & \textbf{330} & \textbf{2,200}  \\ \hline
\end{tabular}
}
\end{table}
\begin{table}[H]
\caption{Performance comparison of \ours{} vs SOTA SVD-LLM methods on Llama3-8B at different compression ratios (CR). Best results are highlighted with \textbf{bold}.}
\label{tab:main_llama3_detailed}
\resizebox{\textwidth}{!}{%
\renewcommand{\arraystretch}{1.35}
\begin{tabular}{ccccccccccccc}
\hline
                            &                         & \multicolumn{9}{c}{\textbf{Accuracy$\uparrow$}}                                                                           & \multicolumn{2}{c}{\textbf{Perplexity$\downarrow$}}       \\ \cline{3-13}
\multirow{-2}{*}{\textbf{Method}} & \multirow{-2}{*}{\textbf{CR}} & \textbf{PIQA} & \textbf{HellaSwag} & \textbf{LAMBADA} & \textbf{ARC-e} & \textbf{ARC-c} & \textbf{SciQ} & \textbf{Race} & \textbf{MMLU} & \textbf{Avg.} & \textbf{WikiText} & \textbf{LAMBADA} \\ \hline

\textbf{Llama3 8B} & -- & 80.7 & 79.1 & 75.6 & 77.7 & 53.5 & 93.9 & 40.3 & 62.2 & 70.4 & 7.3 & 3.1 \\ \hline

SVD-LLM &  & 71.1 & 58.4 & 59.3 & 55.5 & 34.0 & 86.4 & 35.5 & 32.6 & 54.1 & 41 & 11 \\
CoSpaDi &  & 75.2 & 66.5 & 73.8 & 66.5 & 41.6 & 89.5 & 38.2 & 42.8 & 61.8 & 20 & 4.3 \\
ROCKET  & \multirow{-3}{*}{0.2}
        & \textbf{76.9} & \textbf{74.8} & 73.6 & \textbf{73.7} & \textbf{47.1} & \textbf{92.7} & \textbf{40.7} & \textbf{54.9} & \textbf{66.8} & \textbf{11} & \textbf{3.8} \\ \hline

SVD-LLM &  & 65.8 & 46.4 & 38.1 & 41.9 & 27.7 & 70.0 & 31.8 & 27.2 & 43.6 & 150 & 61 \\
CoSpaDi &  & 70.5 & 56.2 & 61.3 & 54.2 & 33.5 & 85.7 & 36.2 & 32.2 & 53.7 & 45 & 9.2 \\
ROCKET  & \multirow{-3}{*}{0.3}
        & \textbf{76.8} & \textbf{69.5} & \textbf{70.4} & \textbf{70.5} & \textbf{42.3} & \textbf{90.9} & \textbf{39.7} & \textbf{47.8} & \textbf{63.5} & \textbf{15} & \textbf{4.4} \\ \hline

SVD-LLM &  & 60.3 & 34.5 & 11.4 & 32.4 & 24.5 & 44.2 & 25.7 & 23.1 & 32.0 & 550 & 1,300 \\
CoSpaDi &  & 63.7 & 41.4 & 30.3 & 39.1 & 26.6 & 68.5 & 30.5 & 25.4 & 40.7 & 180 & 120 \\
ROCKET  & \multirow{-3}{*}{0.4}
        & \textbf{71.9} & \textbf{60.4} & \textbf{59.3} & \textbf{64.4} & \textbf{36.4} & \textbf{88.1} & \textbf{36.8} & \textbf{39.7} & \textbf{57.1} & \textbf{29} & \textbf{8.1} \\ \hline

SVD-LLM &  & 55.1 & 24.7 & 1.8 & 26.0 & 23.1 & 30.5 & 22.4 & 21.0 & 25.6 & 1,800 & 6,500 \\
CoSpaDi &  & 58.4 & 31.8 & 13.6 & 30.7 & 24.8 & 46.2 & 25.8 & 23.4 & 31.8 & 740 & 520 \\
ROCKET  & \multirow{-3}{*}{0.5}
        &\textbf{66.9}& \textbf{45.6} & \textbf{28.4} & \textbf{48.2} & \textbf{27.7} &\textbf{78.2}  & \textbf{33.2} &\textbf{26.1}  & \textbf{44.3} & \textbf{116} & \textbf{90}  \\ \hline

\end{tabular}
}
\end{table}

\subsection{Scaling Behavior Across Model Sizes}
\label{scaling_property}
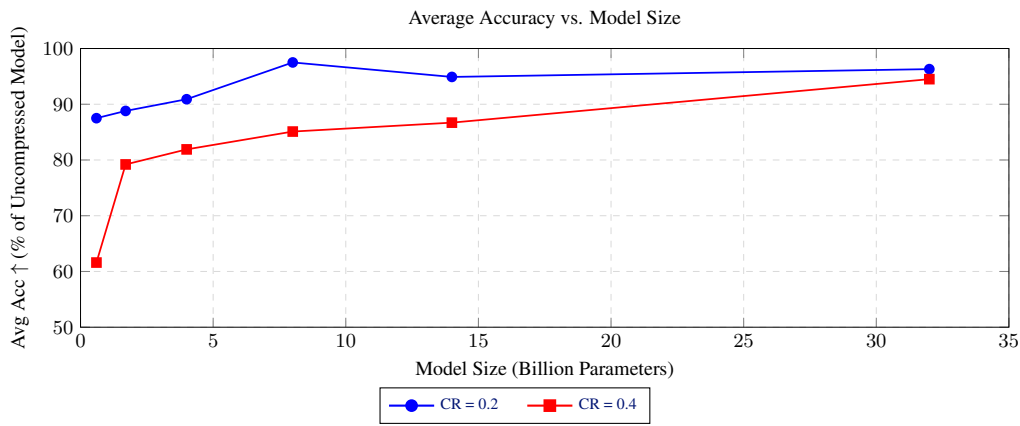
\begin{figure}[H]
  \centering
  \small
  \resizebox{0.8\textwidth}{!}{
  \begin{tikzpicture}
    \begin{axis}[
      width=0.95\textwidth, height=6cm,
      title={Average Accuracy vs. Model Size},
      xlabel={Model Size (Billion Parameters)},
      ylabel={Avg Acc $\uparrow$ (\% of Uncompressed Model)},
      xmin=0, xmax=35,
      ymin=50, ymax=100,
      xtick={0,5,10,15,20,25,30,35},
      ytick={50,60,...,100},
      grid=major,
      grid style={dashed, gray!30},
      legend to name=sharedlegend_acc,
      legend columns=2,
      legend style={font=\tiny, draw=black, fill=white, /tikz/every even column/.append style={column sep=0.3cm}},
      legend cell align={left},
    ]
      % CR = 0.2
      \addplot[blue, mark=*, thick] coordinates {
        (0.6, 87.5)
        (1.7, 88.8)
        (4,   90.9)
        (8,   97.5)
        (14,  94.9)
        (32,  96.3)
      };
      \addlegendentry{CR = 0.2}

      % CR = 0.4
      \addplot[red, mark=square*, thick] coordinates {
        (0.6, 61.6)
        (1.7, 79.2)
        (4,   81.9)
        (8,   85.1)
        (14,  86.7)
        (32,  94.5)
      };
      \addlegendentry{CR = 0.4}
    \end{axis}
  \end{tikzpicture}
  }
  \vspace{2ex}
  \ref{sharedlegend_acc}
  \caption{Average accuracy (relative to uncompressed model) as a function of model size for two compression ratios.}
  \label{fig:acc_vs_model_size}
\end{figure}
In Figure~\ref{fig:acc_vs_model_size}, we present the results of compressing Qwen models of varying sizes from 0.6B to 32B parameters at two compression ratios (20\% and 40\%). The results show that larger models retain a higher fraction of their original (uncompressed) performance after compression. This suggests that larger models may still be significantly underfitted relative to their capacity.
\subsection{Evaluation on Advanced Benchmarks}

\begin{table}[H]
\centering
\caption{Performance comparison of \ours{} against CoSpaDi and SVD-LLM on a new set of benchmarks.}
\label{tab:newlb}
\resizebox{0.7\textwidth}{!}{%
\renewcommand{\arraystretch}{1.35}
\begin{tabular}{l c ccccccc}
\hline
                            &                          & \multicolumn{6}{c}{\textbf{Accuracy$\uparrow$}} \\
\cline{3-8}
\multirow{-2}{*}{\textbf{Method}} & \multirow{-2}{*}{\textbf{CR}} & \textbf{IFEval} & \textbf{BBH} & \textbf{MATH} & \textbf{GPQA} & \textbf{MUSR} & \textbf{MMLU-Pro} \\
\hline
Qwen3 8B                    & —   & 39.21 & 60.86 & 52.57 & 36.16 & 43.12 & 47.72 \\
\hline
SVD-LLM                     & 0.2 & 25.54 & 41.00 & 1.06  & 28.36 & 39.81 & 26.30 \\
CoSpaDi                     & 0.2 & 28.90 & 45.25 & 1.96  & 28.61 & \textbf{42.06} & 31.46 \\
\textbf{ROCKET}             & \textbf{0.2} & \textbf{31.89} & \textbf{50.33} & \textbf{11.10} & \textbf{31.45} & 39.68 & \textbf{36.65} \\
\hline
SVD-LLM                     & 0.3 & 22.90 & 34.42 & 0.98  & 25.59 & \textbf{41.40} & 18.82 \\
CoSpaDi                     & 0.3 & 25.18 & 38.22 & 0.98  & 24.75 & 38.36 & 22.81 \\
\textbf{ROCKET}             & \textbf{0.3} & \textbf{25.54} & \textbf{47.54} & \textbf{2.64} & \textbf{29.19} & 39.94 & \textbf{32.27} \\
\hline
SVD-LLM                     & 0.4 & 22.66 & 30.24 & 0.83  & 23.07 & 37.70 & 11.55 \\
CoSpaDi                     & 0.4 & \textbf{26.14} & 32.72 & 0.76  & 26.01 & 38.10 & 16.74 \\
\textbf{ROCKET}             & \textbf{0.4} & 23.99 & \textbf{40.06} & \textbf{1.28} & \textbf{28.85} & \textbf{41.00} & \textbf{25.53} \\
\hline
\end{tabular}%
}
\end{table}
In Table \ref{tab:newlb} we compare against other methods on a new set of benchmarks. These modern benchmarks IFEval \cite{ifeval}, BBH \cite{bbh}, MATH \cite{math}, GPQA \cite{gpqa}, MuSR \cite{musr}, and MMLU-Pro \cite{mmlupro} represent an evolution in the evaluation of large language models (LLMs), targeting more rigorous, diverse, and realistic capabilities. IFEval (Instruction-Following Evaluation) assesses a model’s ability to follow precise, verifiable natural language instructions. BBH (Big-Bench Hard) isolates 23 of the most challenging tasks from the original BIG-Bench suite to probe complex reasoning. MATH evaluates advanced mathematical problem-solving across algebra, geometry, and other domains. GPQA tests graduate-level scientific knowledge with high difficulty and minimal data contamination risk. MuSR (Multi-step Reasoning) focuses on multi-hop and long-context reasoning, while MMLU-Pro enhances the original MMLU by increasing answer choices (from 4 to 10) and reducing ambiguity, thereby offering a cleaner, more demanding assessment of expert knowledge. Unfortunately other papers are still following the old benchmarks therefore for fair comparison we stick with them for comparison in the main paper.

\subsection{Cross-Architecture and Large-Scale Evaluations}
\label{subsec:cross_arch_eval}
To validate cross-architecture generality, we evaluate \ours{} on Gemma3-12B, Phi-4-14B, Qwen3-30B-A3B-MoE, and Llama3-70B. Results are summarized in Table~\ref{tab:cross_arch_results}.

\begin{table}[h]
\centering
\caption{ROCKET performance across diverse LLM families and scales.}
\label{tab:cross_arch_results}
\resizebox{0.7\textwidth}{!}{%
\begin{tabular}{lccc}
\toprule
\textbf{Model} & \textbf{CR} & \textbf{Avg. Accuracy ($\uparrow$)} & \textbf{Lambada PPL ($\downarrow$)}  \\
\midrule
Gemma3-12B (dense) & -- & 72.88 & 4.16  \\
SVD-LLM & 0.3 & 51.70 & 74.61  \\
\ours{} & 0.3 & \textbf{68.50} & \textbf{4.23}  \\
\ours{} & 0.4 & \textbf{62.75} & \textbf{8.36} \\ \midrule
Phi-4-14B (dense) & -- & 72.05 & 3.49  \\
SVD-LLM & 0.3 & 64.62 & 4.39  \\
\ours{} & 0.3 & \textbf{72.39} & \textbf{3.06} \\
\ours{} & 0.4 & \textbf{68.20} & \textbf{3.67} \\ \midrule
Qwen3-30B-A3B-MoE & -- & 72.90 & 4.13  \\
SVD-LLM & 0.3 & 46.12 & 262.19  \\
\ours{} & 0.3 & \textbf{67.90} & \textbf{13.70}  \\
\ours{} & 0.4 & \textbf{54.80} & \textbf{107.30} \\ \midrule
Llama3-70B (dense) & -- & 77.30 & 2.69  \\
SVD-LLM & 0.2 & 72.03 & 2.92  \\
\ours{} & 0.2 & \textbf{76.90} & \textbf{2.71}  \\
SVD-LLM & 0.4 & 55.87 & 11.78 \\
\ours{} & 0.4 & \textbf{73.32} & \textbf{2.54} \\
\bottomrule
\end{tabular}%
}
\end{table}

We also compare \ours{} against structured depth/width pruning methods on Llama1-7B at 20\% compression (Table~\ref{tab:pruning_comparison}). Despite being entirely training-free, \ours{} achieves state-of-the-art average accuracy, outperforming methods that require costly fine-tuning recovery (e.g., LLM-Pruner, LoRAShear).

\begin{table}[h]
\centering
\caption{Comparison with depth/width pruning methods on Llama1-7B (20\% CR).}
\label{tab:pruning_comparison}
\resizebox{\textwidth}{!}{
\begin{tabular}{lcccccccccc}
\toprule
\textbf{Method} & \textbf{Training-free} & \textbf{BoolQ} & \textbf{PIQA} & \textbf{HellaSwag} & \textbf{WinoGrande} & \textbf{ARC-e} & \textbf{ARC-c} & \textbf{OBQA} & \textbf{Avg.} \\
\midrule
Baseline & -- & 76.5 & 79.8 & 76.1 & 70.1 & 72.8 & 47.6 & 57.2 & 68.59 \\
LLM-Pruner & \xmark & 66.79 & 77.58 & 68.48 & 64.96 & 64.06 & 37.88 & 39.00 & 59.82 \\
LoRAPrune & \xmark & 65.82 & 79.31 & 70.00 & 62.76 & 65.87 & 37.69 & 39.14 & 60.05 \\
WANDA & \cmark & 65.75 & 74.70 & 64.52 & 59.35 & 60.65 & 36.26 & 39.40 & 57.23 \\
ShortGPT & \xmark & 68.26 & 72.28 & 61.70 & 63.77 & 60.22 & 39.00 & 41.60 & 58.12 \\
LoRAShear$^\dagger$ & \xmark & 70.17 & 76.89 & 68.69 & 65.83 & 64.11 & 38.77 & 39.97 & 60.63 \\
LoRAShear & \xmark & 72.78 & 76.36 & 69.49 & 67.63 & 69.02 & 39.47 & 40.78 & 62.22 \\
\textbf{ROCKET} & \cmark & \textbf{75.69} & \textbf{78.40} & \textbf{71.80} & \textbf{68.67} & \textbf{71.80} & \textbf{39.51} & \textbf{42.60} & \textbf{64.07} \\
\bottomrule
\end{tabular}
}

\end{table}

\subsection{Post-Compression Quantization}
\label{app:quantization}
Dobi-SVD's reported compression ratios include quantization. To compare at matched effective compression ratios (Table~\ref{tab:llama3_results}), we apply a simple weight-only post-training quantization to \ours{}: both factors $\m F$ and $\m H$ are quantized independently to INT8 with min-max scaling, without calibration data. This conservative setting matches Dobi-SVD's precision and is not tuned for \ours{}. \ours{} is also compatible with stronger quantizers: applying INT8 GPTQ (via LLMCompressor) to \ours{} at 20\% compression on Llama3-8B yields an effective compression of $\approx$60\% with almost no additional degradation (Table~\ref{tab:gptq}).
\begin{table}[h]
\centering
\caption{\ours{} combined with INT8 GPTQ on Llama3-8B.}
\label{tab:gptq}
\begin{tabular}{lccc}
\toprule
\textbf{Method} & \textbf{Effective CR} & \textbf{Avg. Acc. $\uparrow$} & \textbf{PPL $\downarrow$} \\
\midrule
\ours{} (20\% CR) & 0.20 & 66.8 & 11.04 \\
\ours{} (20\% CR) + INT8 GPTQ & $\approx$0.60 & 66.6 & 11.04 \\
\bottomrule
\end{tabular}
\end{table}

\subsection{Post-Compression Healing}
\label{healing_detailed}
In Table~\ref{tab:healing_details_v1}, we provide the per-benchmark results corresponding to the healing experiment. As noted, we are approaching the performance of models trained from scratch and in some cases nearly matching or surpassing them despite using a compressed model. Specifically, we compressed Qwen3-14B to an 8B model and applied a very limited fine-tuning phase (using only approximately 30 million tokens) while keeping the sparsity pattern fixed an approach that is known to be suboptimal. Nevertheless, the resulting model achieves performance comparable to the original Qwen3-8B trained from scratch. We expect that fine-tuning with higher quality, carefully curated data would further improve results. Moreover, as previously mentioned, enabling trainable sparsity patterns remains a direction for future work. 

\begin{table*}[h]
\centering
\caption{Post-compression healing results on Qwen3-8B models.}
\label{tab:healing_details_v1}
\resizebox{\textwidth}{!}{%
\begin{tabular}{lccccccccccl}
\toprule
\textbf{Method} & \textbf{PIQA} & \textbf{HellaSwag} & \textbf{Lambada} & \textbf{ARC-e} & \textbf{ARC-c} & \textbf{SciQ} & \textbf{Race} & \textbf{MMLU} & \textbf{WikiText} & \textbf{Perplexity} & \textbf{Avg. Acc.} \\
\midrule
Qwen3-14B (dense) & 79.86 & 78.85 & 67.88 & 82.82 & 60.23 & 96.50 & 43.25 & 77.20 & 11 & 3.7 & 73.32 \\
\midrule
Qwen3-8B (dense)   & 77.70 & \textbf{74.90} & 64.10 & \textbf{80.70} & \textbf{56.70} & \textbf{95.70} & \textbf{40.90} & \textbf{73.00} & \textbf{12} & 4.6 & \textbf{70.46} \\
ROCKET-Qwen3-8B (training-free) & 72.68 & 62.63 & \textbf{70.26} & 67.76 & 44.19 & 91.20 & 39.80 & 59.99 & 25 & \textbf{3.8} & 63.56 \\
ROCKET-Qwen3-8B (healed) & \textbf{78.51} & 74.67 & 65.55 & 75.29 & 53.07 & 93.50 & 37.89 & 65.23 & 13 & 4.7 & 67.96 \\
\bottomrule
\end{tabular}
}
\end{table*}

\section{Further Ablations}
\label{furtherablations}
\subsection{Ablation on calibration data}
\label{data_ablation}
\begin{table}[H]
\caption{Ablation on calibration data for Llama3.2-1B at 20\% compression (CR = 0.2). The first row shows the uncompressed baseline. Higher accuracy and lower perplexity are better.}
\label{tab:calib_ablation}
\resizebox{\textwidth}{!}{%
\renewcommand{\arraystretch}{1.35}
\begin{tabular}{l c cccccccccc cc}
\hline
                            &                          & \multicolumn{9}{c}{\textbf{Accuracy$\uparrow$}}                                                                                     & \multicolumn{2}{c}{\textbf{Perplexity$\downarrow$}}       \\ \cline{3-13}
\multirow{-2}{*}{\textbf{Calibration Data}} & \multirow{-2}{*}{\textbf{CR}} & \textbf{PIQA} & \textbf{HellaSwag} & \textbf{LAMBADA} & \textbf{ARC-e} & \textbf{ARC-c} & \textbf{SciQ} & \textbf{Race} & \textbf{MMLU} & \textbf{Avg. Acc.} & \textbf{WikiText$_\text{word}$} & \textbf{LAMBADA$_\text{PPL}$} \\
\hline
None (Baseline)   & 0 & 74.5 & 63.7 & 63.0 & 60.5 & 36.2 & 88.3 & 37.8 & 37.0 & 57.6 & 12 & 5.7 \\
\hline
FineWeb        & 0.2 & 71.9 & 56.7 & 46.0 & 56.7 & 32.4 & \textbf{88.6} & \textbf{36.4} & \textbf{30.8} & 52.4 & 18 & 13 \\
PTB               & 0.2 & 70.3 & 54.9 & 47.3 & 55.2 & 31.9 & 87.8 & 34.5 & 26.7 & 51.1 & 21     & 13 \\
WikiText          & 0.2 & 70.7 & 56.3 & 50.9 & \textbf{57.9} & 33.6 & 88.1 & 35.5 & 28.3 & 52.7 & \textbf{17}     & 11 \\
Alpaca            & 0.2 & \textbf{73.7} & \textbf{57.2} & \textbf{53.2} & \textbf{57.9} & \textbf{34.9} & 87.5 & 35.4 & 30.5 & \textbf{53.8} & 20     & \textbf{9.7} \\
\hline
\end{tabular}%
}
\end{table}

% [camera-ready] duplicate compact version of Table tab:calib_ablation (same label) commented out
% \begin{table}[h]
% \caption{Ablation on calibration data for Llama3-1B at 20\% compression.}
% \label{tab:calib_ablation}
% \centering
% \renewcommand{\arraystretch}{1.0}
% \begin{tabular}{l c c}
% \toprule
% \textbf{Calibration Data} & \textbf{Avg. Acc.$\uparrow$} & \textbf{Perplexity$\downarrow$} \\
% \hline
% None (Baseline) & 57.6 & 1.2E+01 \\
% \hline
% PTB     & 51.1 & 2.1E+01 \\
% WikiText   & 52.7 & \textbf{1.7E+01} \\
% Alpaca   & \textbf{53.8} & 2.0E+01 \\
% RefinedWeb (ours) & 52.4 & 1.8E+01 \\
% \hline
% \end{tabular}
% \end{table}

We evaluate the sensitivity of ROCKET to the choice of calibration dataset by comparing four sources: FineWeb~\cite{refinedweb}, PTB~\cite{ptb}, WikiText~\cite{wikitext}, and Alpaca~\cite{alpaca}. As shown in Table~\ref{tab:calib_ablation}, while instruction-tuned data such as Alpaca yield slightly higher average accuracy (53.8 vs.\ 52.4), the differences across datasets are relatively small, confirming that ROCKET is robust to the calibration data choice. Nevertheless, to ensure a fair comparison with CoSpaDi, which uses FineWeb as its default calibration data, we adopt FineWeb for all primary experiments reported in this paper.

\subsection{Ablation on the Sparsification Strategy}
\label{app:sparsity_strategy_details}

\begin{table}[H]
\caption{Ablation on sparsification strategies for Llama3.2-1B at 20\% compression (CR = 0.2). All methods use the same calibration data and global parameter budget.}
\label{tab:sparsity_ablation}
\resizebox{\textwidth}{!}{%
\renewcommand{\arraystretch}{1.35}
\begin{tabular}{l c cccccccccc cc}
\hline
                            &                          & \multicolumn{9}{c}{\textbf{Accuracy$\uparrow$}}                                                                                     & \multicolumn{2}{c}{\textbf{Perplexity$\downarrow$}}       \\ \cline{3-13}
\multirow{-2}{*}{\textbf{Sparsification Strategy}} & \multirow{-2}{*}{\textbf{CR}} & \textbf{PIQA} & \textbf{HellaSwag} & \textbf{LAMBADA} & \textbf{ARC-e} & \textbf{ARC-c} & \textbf{SciQ} & \textbf{Race} & \textbf{MMLU} & \textbf{Avg. Acc.} & \textbf{WikiText$_\text{word}$} & \textbf{LAMBADA$_\text{PPL}$} \\
\hline
None (Baseline)   & 0 & 74.5 & 63.7 & 63.0 & 60.5 & 36.2 & 88.3 & 37.8 & 37.0 & 57.6 & 12 & 5.7 \\
\hline
ROCKET            & 0.2 & \textbf{71.9} & \textbf{56.7} & 46.0 & \textbf{56.7} & 32.4 & \textbf{88.6} & \textbf{36.4} & \textbf{30.8} & \textbf{52.4} & \textbf{18} & \textbf{13} \\
Per-Row Sparsification & 0.2 & 67.8 & 48.1 & 32.5 & 48.6 & 29.4 & 79.9 & 31.6 & 26.7 & 45.6 & 32 & 38 \\
Global Importance Sparsification & 0.2 & 69.5 & 54.7 & 42.7 & 53.5 & 30.7 & 85.7 & 35.5 & 28.1 & 50.1 & 19 & 17 \\
Whitened-Space Importance Only & 0.2 & 71.5 & 55.7 & \textbf{47.3} & 56.3 & \textbf{32.9} & 87.9 & 35.4 & 28.9 & 52.0 & 19 & \textbf{13} \\
\hline
\end{tabular}%
}
\end{table}

To evaluate the impact of our sparsification strategy, we compare \ours{} against three alternative approaches for pruning the coefficient matrix $\m C = \bm\Sigma \m V^\top$: 
(1) \textbf{Per-Row Sparsification}, where importance scores are computed identically but sparsity is enforced independently per row (breaking column-wise structure); 
(2) \textbf{Global Importance Sparsification}, which ignores any structural constraints and simply zeros out the globally least-important entries based on the full importance matrix; and 
(3) \textbf{Whitened-Space Importance Only}, which disables the original-space fidelity term by setting $\lambda = 0$ in~\eqref{eq:imp_final} (theoretically optimal with respect to the whitened space). 
Table~\ref{tab:sparsity_ablation} shows that \ours{}'s \textbf{column-aware, activation-and-weight-balanced sparsification} outperforms alternatives, demonstrating the efficiency of both structural awareness and dual-space importance scoring.

\subsection{Search Space Construction}
\label{app:search_space}
To maintain tractability, we decouple the sparsity ratio ($k/s$) from the dynamic programming (DP) search space. For each candidate $cr$, we greedily select the $k/s$ pair that minimizes reconstruction error, leaving the DP solver to optimize only over the discretized $cr$ candidates. In practice, we use a \textbf{$cr$ step of 0.05} and restrict \textbf{$k/s \in [1, 4]$} with step 0.1, yielding 14 compression candidates per layer, each evaluated across 30 sparsity configurations. Crucially, DP complexity depends on the number of layers and candidates (Appendix~\ref{sec:scalability_runtime}), \textit{not} on weight matrix dimensions.
\subsection{Ridge Regularization of the Dictionary Update}
\label{hyp_ablation_details}
When solving for the refined left factor $\m{D}$ post-sparsification, we apply a small adaptive ridge regularization $\mu$ to prevent ill-conditioning in $\m{C}\m{C}^\top$:
\[
\mu = \max\left(10^{-6},\ 10^{-3} \cdot \frac{\mathrm{trace}(\m{C}^\top \m{C})}{k}\right).
\]
This keeps regularization negligible relative to matrix scale while ensuring numerical stability during least-squares inversion.

\subsection{Ablation on Reconstruction Error Metric}
\label{app:recontruvtion_error_ablations}

In ROCKET, the layer profiling stage enumerates a set of candidate compression configurations per layer, where each candidate is defined by a pair $(cr, ks)$, namely, the compression ratio and the sparsity ratio applied to the coefficient matrix. Given the original layer weights $\m W$, each candidate $(cr, ks)$ induces a reconstructed weight matrix $\tilde{\m W}$. For each candidate, we compute a reconstruction error, which serves as the estimated \emph{cost} in the constrained multi-choice knapsack problem. The optimizer then selects one candidate per layer to minimize the sum of estimated error while satisfying the global parameter budget.

We evaluate four variants for this per-candidate error estimate: (1) \textbf{relative Frobenius error} $\|\m W - \tilde{\m W}\|_F / \|\m W\|_F$ (our default), (2) \textbf{$\ell_1$ distance} $\|\m W - \tilde{\m W}\|_1$, (3) \textbf{mean cosine distance across columns}, and (4) \textbf{spectral distance} $\|\m W - \tilde{\m W}\|_2$. All variants run under the same configurations. As shown in Table~\ref{tab:error_metric_ablation}, the relative Frobenius error yields the best downstream performance, while $\ell_1$-based estimates lead to significant degradation, highlighting the effectiveness of the former metric for effective budget allocation.

\begin{table}[h]
    \centering
    \caption{Ablation on reconstruction error metric for layer profiling in ROCKET for Llama3.2-1B at 20\% compression.}
    \label{tab:error_metric_ablation}
    \renewcommand{\arraystretch}{1.0}
    \begin{tabular}{l c c}
    \toprule
    \textbf{Error Metric} & \textbf{Avg. Acc. $\uparrow$} & \textbf{Perplexity $\downarrow$} \\
    \midrule
    Dense (Baseline)        & 57.6 & 12 \\
    \midrule
    L1 Distance            & 35.2 & 180 \\
    Mean Cos Columns       & 51.1 & 19 \\
    Spectral Distance      & 51.3 & 18 \\
    Frobenius (ours)       & \textbf{52.4} & \textbf{18} \\
    \bottomrule
    \end{tabular}
\end{table}

\end{document}